\documentclass{article}

\usepackage{microtype}
\usepackage{graphicx}
\usepackage{subfigure}
\usepackage{booktabs} % for professional tables

\usepackage{microtype}
\usepackage{graphicx}
\usepackage{amsmath}
\usepackage{adjustbox}
\usepackage{float}
\usepackage{xcolor}
\usepackage{amssymb}
\usepackage{bm}
\usepackage{multirow}
\usepackage[ruled,vlined]{algorithm2e}
\usepackage{placeins}

\usepackage{tabularx}
\newcolumntype{L}{>{\raggedright\arraybackslash}X}
\newcolumntype{Y}{>{\centering\arraybackslash}X}

\usepackage{hyperref}

\usepackage[accepted]{icml2020}

\icmltitlerunning{Explore, Discover and Learn: Unsupervised Discovery of State-Covering Skills}

\begin{document}

\twocolumn[
\icmltitle{Explore, Discover and Learn: \\Unsupervised Discovery of State-Covering Skills}



\icmlsetsymbol{equal}{*}

% \begin{icmlauthorlist}
% \icmlauthor{Anonymous}{anonymous}
% \end{icmlauthorlist}

% \icmlaffiliation{anonymous}{Anonymous}

% \icmlcorrespondingauthor{Anonymous}{anonymous@anonymous.com}

\begin{icmlauthorlist}
\icmlauthor{V\'{\i}ctor Campos}{bsc}
\icmlauthor{Alex Trott}{salesforce}
\icmlauthor{Caiming Xiong}{salesforce}
\icmlauthor{Richard Socher}{salesforce}
\icmlauthor{Xavier Giro-i-Nieto}{upc}
\icmlauthor{Jordi Torres}{bsc}
\end{icmlauthorlist}

\icmlaffiliation{bsc}{Barcelona Supercomputing Center}
\icmlaffiliation{salesforce}{Salesforce Research}
\icmlaffiliation{upc}{Universitat Polit\`{e}cnica de Catalunya}

\icmlcorrespondingauthor{V\'{\i}ctor Campos}{victor.campos@bsc.es}

\icmlkeywords{Machine Learning, ICML}

\vskip 0.3in
]



\printAffiliationsAndNotice{}  % leave blank if no need to mention equal contribution

\begin{abstract}

Acquiring abilities in the absence of a task-oriented reward function is at the frontier of reinforcement learning research. 
This problem has been studied through the lens of \textit{empowerment}, which draws a connection between option discovery and information theory.
Information-theoretic skill discovery methods have garnered much interest from the community, but little research has been conducted in understanding their limitations. 
Through theoretical analysis and empirical evidence, we show that existing algorithms suffer from a common limitation -- they discover options that provide a poor coverage of the state space. 
In light of this, we propose \textit{Explore, Discover and Learn} (EDL), an alternative approach to information-theoretic skill discovery. 
Crucially, EDL optimizes the same information-theoretic objective derived from the empowerment literature, but addresses the optimization problem using different machinery.
We perform an extensive evaluation of skill discovery methods on controlled environments and show that EDL offers significant advantages, such as overcoming the coverage problem,
reducing the dependence of learned skills on the initial state,
% reducing the dependence of learned skills on the state distribution experienced during early training,
and allowing the user to define a prior over which behaviors should be learned.
Code is publicly available at \url{https://github.com/victorcampos7/edl}.

\end{abstract}

\section{Introduction}

\begin{figure}[t]
    \centering
    \resizebox{\linewidth}{!}{\includegraphics{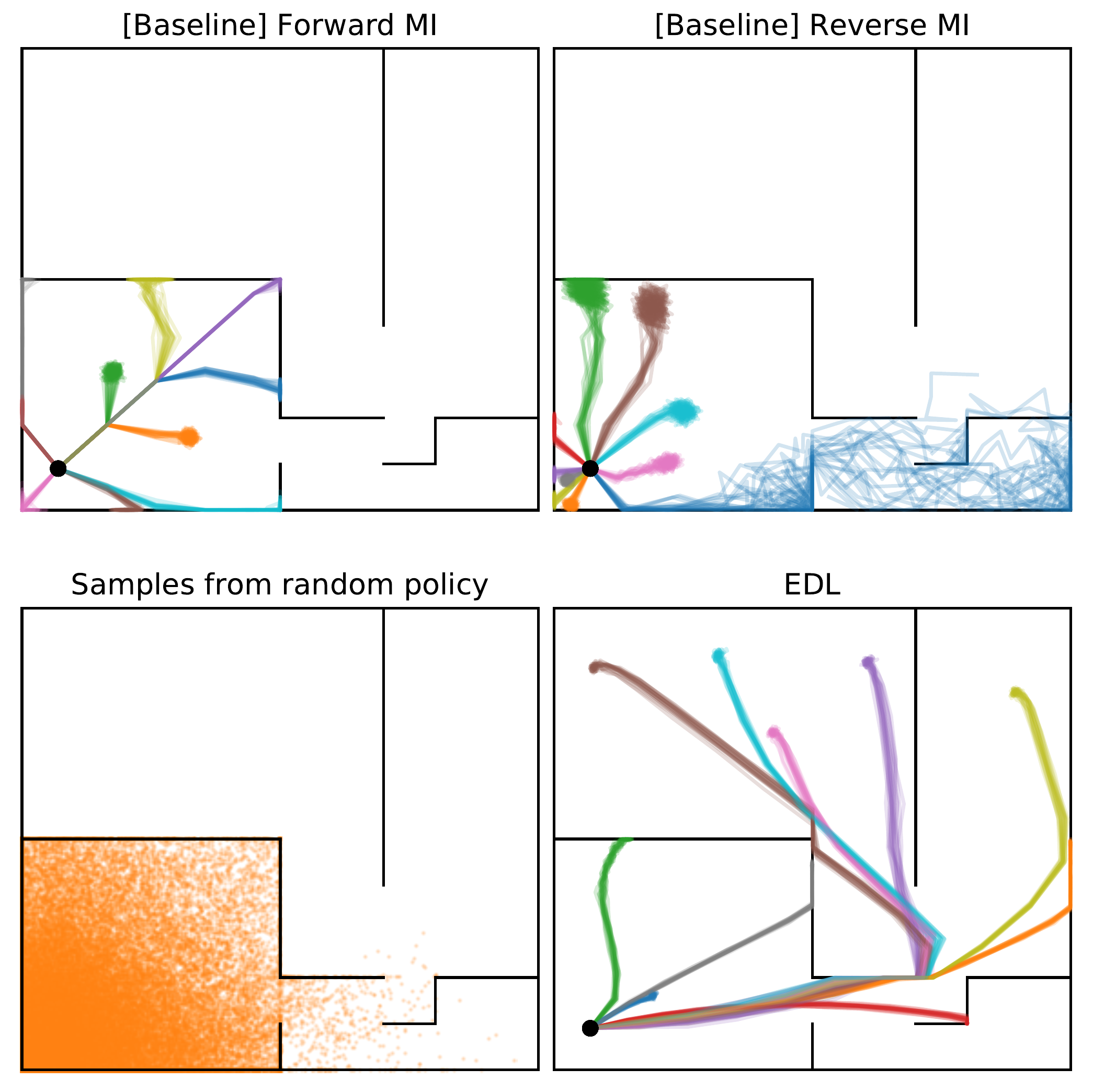}}
    \caption{Skills learned on a maze with bottleneck states. Each colored line represents a trajectory initiated at the black dot by a different skill. Multiple rollouts per skill are reported in order to account for the stochasticity of the policy.
    % Each skill is shown in a different color, and 20 rollouts are sampled per skill in order to account for the stochasticity of the policy. All trajectories start at the position marked by the black dot. 
    % Existing methods fail at exploring beyond the bottleneck state, whereas the proposed approach discovers skills that provide a better coverage of the state space.
    The bottom left plot depicts states visited by a policy with random weights, showing which states are reachable by the agent at the beginning of training.
    Existing methods fail at expanding this set of states, and end up committing to behaviors discovered by the random policy. 
    On the other hand, EDL discovers skills that provide a better coverage of the state space.
    }
    \label{fig:motivation}
    \label{fig:bottleneck_maze}
\end{figure}

Reinforcement learning (RL) algorithms have recently achieved outstanding goals thanks to advances in simulation~\cite{todorov2012_mujoco,bellemare2013_ale}, efficient and scalable learning algorithms~\cite{mnih2015_dqn,lillicrap2016_ddpg,schulman2015_trpo,schulman2017_ppo,espeholt2018_impala}, function approximation~\cite{lecun2015_deeplearning,goodfellow2016_deeplearningbook}, and hardware accelerators~\cite{nvidia2017_v100,jouppi2017_tpu}. These landmarks include outperforming humans in board~\cite{silver2017_alphazero} and computer games~\cite{mnih2015_dqn,vinyals2019_alphastar}, and solving complex robotic control tasks~\cite{andrychowicz2017_her,akkaya2019_rubik}. 
% All these problems have one property in common -- the existence of a reward function that measures progress and drives learning. 
% However, designing reward functions that guide agents towards the desired behavior is a challenging problem~\cite{hadfield2017_inverse,lehman2011_novelty}. Even when such a function exists, the cost of evaluating it on a large number of samples might preclude its usage, e.g.~when human feedback is required~\cite{christiano2017_preferences}.
Typically, training aims to solve a particular task, relying on task-specific reward functions to measure progress and drive learning. This contrasts with how intelligent creatures learn in the absence of external supervisory signals, acquiring abilities in a task-agnostic manner by exploring the environment. 
Methods for training models without expert supervision have already obtained promising results in fields like natural language processing~\cite{radford2019_gpt2,devlin2019_bert} and computer vision~\cite{henaff2019_cpc,he2019_moco}.
In RL, analogous ``unsupervised'' methods are often aimed at learning generically useful behaviors for interacting within some environment, behaviors that may naturally accelerate learning once one or more downstream tasks become available.

% Agents trained in this unsupervised fashion hold the promise of acquiring generic knowledge that can be used to learn new tasks quickly and efficiently. This is a treasured property in RL, with current methods usually suffering from sample inefficiency.

% The difficulty of designing reward functions is analogous to that of collecting large-scale annotated datasets for supervised methods. The latter has motivated the development of techniques that leverage large collections of unlabeled data, finding great success in fields such as natural language processing~\cite{radford2019_gpt2,devlin2019_bert} and computer vision~\cite{henaff2019_cpc,he2019_moco}. These methods are based on self-supervision, i.e.~they generate a supervisory signal automatically from input data and without making use of task-specific annotations. In the RL framework, this unsupervised setup amounts to automatically inferring a task-agnostic reward signal that can be used to drive training of agents.

The idea of unsupervised RL is often formulated through the lens of \textit{empowerment}~\cite{salge2014_empowerment}, which formalizes the notion of an agent discovering \textit{what} can be done in an environment while learning \textit{how} to do it. Central to this formulation is the concept of mutual information, a tool from information theory. 
\citet{mohamed2015_variational} derived a variational lower bound on the mutual information which can be used to learn options~\cite{sutton1999_between} in a task-agnostic fashion. Following classical empowerment~\cite{salge2014_empowerment}, options are discovered by maximizing the mutual information between sequences of actions and final states. This results in \textit{open loop} options, where the agent commits to a sequence of actions \textit{a priori} and follows them regardless of the observations received from the environment.
\citet{gregor2016_vic} developed an algorithm to learn \textit{closed loop} options, whose actions are conditioned on the state, by maximizing the mutual information between states and some latent variables instead of action sequences.
This approach has been extended by several works, which are surveyed in Section~\ref{sec:mi_based_methods}. 
Despite the interest in developing information-theoretic skill discovery methods, very little research has been conducted in understanding the limitations of these algorithms. We distinguish two categories of such limitations. The first type has to do with the nature of the objective itself, e.g.~the difficulty of purely information-theoretic methods for capturing human priors~\cite{achiam2018_valor}. In this work, we focus on the second group -- those issues introduced when adapting the objective to current optimization methods. 

In order to maximize empowerment, an agent needs to learn to control the environment while discovering available options. It should not aim for states where it has the most control according to its current abilities, but for states where it expects it will achieve the most control \textit{after} learning~\cite{gregor2016_vic}. We empirically observe that this is not achieved by existing methods, which prematurely commit to already discovered options instead of exploring the environment to unveil novel ones. Figure~\ref{fig:motivation} (top) showcases this failure mode when deploying existing algorithms on a 2D maze. 
% Our analysis suggests that this might be caused by the generative model for states and skills considered in these methods. 
We provide theoretical analysis showing that these methods tend to reinforce already discovered behaviors at the expense of exploring in order to discover new ones, resulting in behaviors that exhibit poor coverage of the available state space.
Figure~\ref{fig:motivation} (bottom right) depicts the skills discovered by our proposed \textit{Explore, Discover and Learn} (EDL) paradigm, a three-stage methodology that is able to discover skills with much better coverage.

Our contributions can be summarized as follows. 
(1)~We provide theoretical analysis and empirical evidence showing that existing skill discovery algorithms fail at learning state-covering skills. 
(2)~We propose an alternative approach to information-theoretic option discovery, \textit{Explore, Discover and Learn} (EDL), that overcomes the limitations of existing methods. Crucially, EDL achieves this while optimizing the same information-theoretic objective as previous methods.
(3)~We validate the presented paradigm by implementing a solution that follows the three-stage methodology. Through extensive evaluation in controlled environments, we demonstrate the effectiveness of EDL, showcase its advantages over existing methods, and analyze its current limitations and directions for future research.

\section{Information-theoretic skill discovery}
\label{sec:mi_based_methods}

This section presents a generic mathematical framework that can be used to formulate, analyze and compare information-theoretic skill discovery methods in the literature.
Let us consider a Markov Decision Process (MDP) $\mathcal{M} \equiv (\mathcal{S}, \mathcal{A}, p)$ with state space $\mathcal{S}$, action space $\mathcal{A}$ and transition dynamics $p$. 
We learn latent-conditioned policies $\pi(a|s,z)$, and define \textit{skills} or \textit{options} as the policies obtained when conditioning $\pi$ on a fixed value of $z \in \mathcal{Z}$~\cite{eysenbach2019_diayn,achiam2018_valor}.
Let $S \sim p(s)$ be a random variable denoting states such that $S \in \mathcal{S}$, and $Z \sim p(z)$ be a random variable for latent variables. Using notation from information theory, we use $I(\cdot ; \cdot)$ and $H(\cdot)$ to refer to the mutual information and Shannon entropy, respectively. Information-theoretic skill discovery methods seek to find a policy that maximizes the mutual information between $S$ and $Z$. 
% We define \textit{skills} as the policies obtained when conditioning $\pi$ on a fixed value of $z \in \mathcal{Z}$.
Due to symmetry, this measure can be expressed in the following two forms:
\begin{align}
    I(S;Z) 
    &= H(Z) - H(Z|S) \label{eq:reverse_mi}  && \hspace{5pt}\textit{// reverse} \\
    &= H(S) - H(S|Z) \label{eq:forward_mi}  && \hspace{5pt}\textit{// forward}
\end{align}
For presentation clarity, we follow \citet{gregor2016_vic} and refer to Equations \ref{eq:reverse_mi} and \ref{eq:forward_mi} as the \textit{reverse} and \textit{forward} forms of the mutual information, respectively. 
% We classify existing skill discovery methods depending on the form of the mutual information they optimize.

Our goal is to analyze which fundamental design choices are responsible for the properties and limitations of such algorithms. 
We classify existing skill discovery methods depending on the form of the mutual information they optimize, and implement a canonical algorithm for each form that allows for fair comparison.
The following subsections describe existing skill discovery methods as well as the specific implementations considered in this work.

\subsection{Reverse form of the mutual information}
\label{sec:reverse_mi}

The objective can be derived by expanding the definition of the mutual information in Equation~\ref{eq:reverse_mi}, and then leveraging the non-negativity property of the KL divergence to compute a variational lower bound~\cite{barber2003_im}:
\begin{align}
    I(S;Z) &= \mathbb{E}_{s,z \sim p(s,z)} [\log p(z|s)] - \mathbb{E}_{z \sim p(z)}[\log p(z)] \\
    &\geq \mathbb{E}_{s,z \sim p(s,z)} [\log q_{\phi}(z|s)] - \mathbb{E}_{z \sim p(z)}[\log p(z)] \label{eq:reverse_mi_bound}
\end{align}
where $q_{\phi}(z|s)$ is fitted by maximum likelihood on $(s,z)$-tuples collected by deploying the policy in the environment. This implicitly approximates the unknown posterior as $p(z|s) \approx \rho_{\pi}(z|s)$, where $\rho_{\pi}(z|s)$ is the empirical posterior induced by the policy.

Note that computing this measure will require sampling from two distributions, $p(z)$ and $p(s,z)$. The distribution over latent variables $p(z)$ can be learned as part of the optimization process or fixed beforehand, with the latter often yielding superior results~\cite{eysenbach2019_diayn}. However, sampling from the joint distribution over states and latents $p(s,z)$ is more problematic. A common workaround consists in assuming a generative model of the form $p(s,z) = p(z) p(s|z) \approx p(z) \rho_{\pi}(s|z)$, where $\rho_{\pi}(s|z)$ is the stationary state-distribution induced by $\pi(a|s,z)$~\cite{gregor2016_vic}.

This category includes a variety of methods with slight differences. VIC~\cite{gregor2016_vic} considers only the final state of each trajectory and a learnable prior $p(z)$. SNN4HRL~\cite{florensa2017_snn} introduces a task-specific proxy reward, which encourages exploration and can be understood as a bonus to increase the entropy of the stationary state-distribution. DIAYN~\cite{eysenbach2019_diayn} additionally minimizes the mutual information between actions and skills given the state, resulting in a formulation that resembles MaxEnt RL~\cite{haarnoja2017_softdqn}. VALOR~\cite{achiam2018_valor} considers the posterior over sequences of states instead of individual states in order to encourage learning dynamical modes rather than goal-attaining modes. VISR
~\cite{hansen2020_visr} combines skill discovery with universal successor features approximators~\cite{borsa2019_universal} to enable fast task inference~\cite{barreto2017_successor,barreto2018_transfer}. DISCERN~\cite{warde2019_discern} and Skew-Fit~\cite{pong2019_skewfit} aim at learning a goal-conditioned policy in an unsupervised fashion, which can be understood as skill discovery methods where $Z$ takes the form of states sampled from a buffer of previous experience.

We consider a variant of VIC~\cite{gregor2016_vic} with a fixed prior $p(z)$ and where all states in a trajectory are considered in the objective\footnote{The original implementation by \citet{gregor2016_vic} considered final states only, thus providing a sparser reward signal to the policy.}. This method can be seen as a version of DIAYN~\cite{eysenbach2019_diayn} where the scale of the entropy regularizer $H(a|s,z)$ is set to 0. The variational lower bound in Equation~\ref{eq:reverse_mi_bound} is optimized by training the policy $\pi(a|s,z)$ using the reward function
\begin{align}
    r(s,z') = \log q_{\phi}(z'|s) - \log p(z'), \, z' \sim p(z)  \label{eq:reverse_mi_rew}
\end{align}

\subsection{Forward form of the mutual information} 
\label{sec:forward_mi}

A similar lower bound to that in Equation~\ref{eq:reverse_mi_bound} can be derived by expanding the forward form of the mutual information in Equation~\ref{eq:forward_mi}:
\begin{align}
    I(S;Z) &= \mathbb{E}_{s,z \sim p(s,z)} [\log p(s|z)] - \mathbb{E}_{s \sim p(s)}[\log p(s)] \\
    &\geq \mathbb{E}_{s,z \sim p(s,z)} [\log q_{\phi}(s|z)] - \mathbb{E}_{s \sim p(s)}[\log p(s)] \label{eq:forward_mi_bound}
\end{align}
where $q_{\phi}(s|z)$ is fitted by maximum likelihood on $(s,z)$-tuples collected by deploying the policy in the environment. This amounts to approximating $p(s|z)$ with the stationary state-distribution of the policy, $p(s|z) \approx \rho_{\pi}(s|z)$.

To the best of our knowledge, DADS~\cite{sharma2019_dynamics} is the only method within this category. 
DADS follows a model-based setup where $I(S_{t+1};Z|S_t)$ is maximized. This is achieved by modelling changes in the state, $\Delta s = s_{t+1} - s_t$. When evaluated on locomotion environments that encode the position of the agent in the state vector, this setup favors the discovery of gaits that move in different directions.
Similarly to methods in Section~\ref{sec:reverse_mi}, $p(s,z)$ is approximated by relying on the stationary state-distribution induced by the policy and $p(s) \approx \rho_{\pi}(s) = \mathbb{E}_z \left[ \rho_{\pi}(s|z) \right]$. 
We will consider a model-free variant of DADS where the variational lower bound in Equation~\ref{eq:forward_mi_bound} is optimized by training the policy $\pi(a|s,z)$ with a reward function
\begin{align}
    r(s,z') = \log q_{\phi}(s|z') - \log \frac{1}{L} \sum_{i=1}^{L} q_{\phi}(s|z_i), \, z', z_i \sim p(z) \label{eq:forward_mi_rew}
\end{align}
where $p(s)$ is approximated using $q_{\phi}$ and $L$ random samples from the prior $p(z)$ as done by \citet{sharma2019_dynamics}. When using a discrete prior, we marginalize over all skills.

\subsection{Limitations of existing methods}

% Short intro to the problem
Recall that maximizing empowerment implies fulfilling two tasks, namely discovering what is possible in the environment and learning how to achieve it. In preliminary experiments, we observed that existing methods discovered skills that provide a poor coverage of the state space. This suggests a limited capability for discovering what options are available.
%, which is fundamental to becoming empowered. 

% Assumptions in existing methods
Maximizing the mutual information between states and latents requires knowledge of some distributions. Methods based on the forward form of the mutual information make use of $p(s|z)$ and $p(s)$, whereas those using the reverse form employ $p(z|s)$. Note that none of these are known \textit{a priori}, so the common practice is to approximate them using the distributions induced by the policy. Distributions over states are approximated with the stationary state-distribution of the policy, $p(s|z) \approx \rho_{\pi}(s|z)$ and $p(s) \approx \rho_{\pi}(s) = \mathbb{E}_z \left[ \rho_{\pi}(s|z) \right]$. The posterior $p(z|s)$ is approximated with the empirical distribution induced by running the policy, $p(z|s) \approx \rho_{\pi}(z|s)$. In practice, these distributions are estimated via maximum likelihood using rollouts from the policy. 

% Theoretical analysis
We analyze the asymptotic behavior of the reward function for existing methods under the aforementioned approximations through a theoretical lens. The analysis considers an agent aiming to discover $N$ discrete skills, and perfect estimations of all distributions. Our main result shows that the agent receives larger rewards for visiting known states than discovering new ones. Known states can receive a reward of up to $r_{\text{max}} = \log N$. On the other hand, previously unseen states will receive a smaller reward, $r_{\text{new}} = 0$. These observations hold for the forward and reverse forms of the mutual information, and provide theoretical insight for why existing methods do not discover state-covering skills. We refer the reader to the Supplementary Material (SM) for a detailed derivation of the results.

% What makes a state 'known'?
% The fact that the policy will initially act randomly only makes this problem worse, as it will likely visit only a small subset of the state space. Although this set of states might be expanded through exploration mechanisms such as random or noisy actions~\cite{mnih2015_dqn,lillicrap2016_ddpg}, entropic action distributions~\cite{haarnoja2017_softdqn,haarnoja2018_sac}, and parameter noise~\cite{plappert2018_parameter,fortunato2018_noisy}, these techniques might not suffice in complex environments~\cite{ecoffet2019_goexplore,florensa2017_reverse}. 

% Experimental evidence
In order to provide preliminary evidence for this result, we deploy the described algorithms on a 2D maze with bottleneck states (see Section~\ref{sec:experiments} for details on the experimental setup). As shown in Figure~\ref{fig:motivation} (top), existing methods fail at exploring the maze and most options just visit different regions of the initial room. Figure~\ref{fig:motivation} (bottom left) depicts states visited by a policy with the same architecture, but random weights. Note that existing algorithms do not expand the set of states visited by this random policy, but simply identify and reinforce different modes of behavior among them. This observation confirms that existing formulations fail at discovering available options, and motivates our study of alternative methods for option discovery.

\section{Proposed method}
\label{sec:proposed_method}

\begin{table*}[t]
    \centering
    \begin{tabularx}{\linewidth}{llL}
         \toprule
         \textbf{ Assumptions} & \textbf{MI form} & \textbf{Methods} \\
         \midrule
         \multirow{2}{*}{
            \shortstack[l]{
                $p(z)$: fixed \\ 
                $p(z|s) \approx \rho_{\pi}(z|s)$ \\
                $p(s|z) \approx \rho_{\pi}(s|z)$ \\ 
                $p(s) \approx \rho_{\pi}(s) = \mathbb{E}_z \left[ \rho_{\pi}(s|z) \right]$ 
                }
            } & 
            Forward & DADS~\cite{sharma2019_dynamics} \\
            \cmidrule{2-3}
            & \multirow{3}{*}{Reverse} & VIC~\cite{gregor2016_vic}, SNN4HRL~\cite{florensa2017_snn}, DIAYN~\cite{eysenbach2019_diayn}, VALOR~\cite{achiam2018_valor}, DISCERN~\cite{warde2019_discern}, Skew-Fit~\cite{pong2019_skewfit}, VISR~\cite{hansen2020_visr} \\
        \midrule
        % $\, \, p(z)$: prior                          & \multirow{3}{*}{Forward}   & \multirow{3}{*}{EDL (ours)} \\
        % $\, \, p(s)$: fixed                          &                            &    \\
        % $\, \, p(s|z), \, p(z|s)$: modelled with VI  &                            &    \\
        $\, \, p(z), p(s)$: fixed                          & \multirow{2}{*}{Forward}   & \multirow{2}{*}{EDL (ours)} \\
        $\, \, p(s|z), \, p(z|s)$: modelled with VI  &                            &    \\
        \bottomrule
    \end{tabularx}
    \caption{Types of methods depending on the considered generative model and the version of the mutual information (MI) being maximized. 
    Distributions denoted by $\rho$ are induced by running the policy in the environment, whereas $p$ is used for the true and potentially unknown ones. 
    The dependency of existing methods on $\rho_{\pi}(s|z)$ causes pathological training dynamics by letting the agent influence over the states considered in the optimization process.
    EDL relies on a fixed distribution over states $p(s)$ to break this dependency and makes use of variational inference (VI) techniques to model $p(s|z)$ and $p(z|s)$.
    }
    \label{tab:types_of_methods}
\end{table*}

Maximizing the mutual information between states and latent variables requires access to unknown distributions, which existing methods approximate using the distributions induced by the policy. Instead of encouraging the agent to discover available options, this approximation reduces the problem to that of reinforcing already discovered behaviors. Since the policy is initialized randomly at the beginning of training, the discovered options seldom explore further than a random policy. 
% This leads us to study alternative ways of modelling the joint distribution.

We propose an alternative approach, \textit{Explore, Discover and Learn} (EDL), for modelling these unknown distributions and performing option discovery.
Existing methods make use of the state distribution $p(s) \approx \rho_{\pi}(s) = \mathbb{E}_z \left[ \rho_{\pi}(s|z) \right]$, which focuses $p(s)$ around states where the policy receives a high reward.
This dependency contributes to the pathological learning dynamics described above.
To break this dependency, EDL makes use of a \textit{fixed} distribution over states $p(s)$ and is agnostic to the method by which this distribution is discovered/obtained.
% to break the dependency on $\rho_{\pi}(s|z)$, which causes pathological training dynamics by letting the agent influence over the states considered in the optimization process.
For a given distribution over states, EDL makes use of variational inference techniques to model $p(s|z)$ and $p(z|s)$.
As its name suggests, EDL is composed of three stages: (\textit{i})~exploration, (\textit{ii})~skill discovery, and (\textit{iii})~skill learning. 
These can be studied and improved upon independently, and the actual implementation of each stage will depend on the problem being addressed.
The compartmentalization of these facets of the objective, together with the inclusion of a fixed distribution over states, are the key features of EDL.
Table~\ref{tab:types_of_methods} positions this new approach with respect to existing ones.

\textbf{Exploration.} In the absence of any prior knowledge, a reasonable choice for the distribution over states $p(s)$ is a uniform distribution over all $\mathcal{S}$, which will encourage the discovery of state-covering skills. 
% Since sampling from an arbitrary distribution over states is generally a difficult problem, we consider possible approaches to achieve this. 
This stage comes with the challenge of being able to generate or sample from the distribution of states that the learned skills should ultimately cover.
This is generally a difficult problem, for which we consider possible solutions. 
When an oracle is available, it can be queried for samples belonging to the set of valid states. If such an oracle is not available, one can train an exploration policy that induces a uniform distribution over states. Finding these policies is known as the problem of maximum entropy exploration, for which provably efficient algorithms exist under certain conditions~\cite{hazan2019_maxent}. When interested in some particular modes of behavior, one can leverage a more specific state distribution or adopt a non-parametric solution by sampling states from a dataset of extrinsically generated experience~\cite{guss2019_minerl}. Note that unlike approaches in imitation learning~\cite{ho2016_gail}, learning from demonstrations~\cite{hester2017_dqfd,vevcerik2017_ddpgfd}, and learning from play~\cite{lynch2019_play}, EDL does not require access to trajectories or actions emitted by an expert policy.

\textbf{Skill discovery.} 
% The considered formulation requires knowledge of either $p(z|s)$ or $p(s|z)$, which are unknown. 
% We turn to variational inference techniques to model these distributions, and we employ Variational Autoencoders (VAE)~\cite{kingma2014_vae} for this purpose. 
Whereas existing methods sample skill $z \sim p(z)$ directly as an input to the skill-conditioned policy, EDL requires an indirect approach wherein latent skills are inferred from $p(s)$.
More concretely, given a distribution over states, or samples from it, we treat skill discovery as learning to model $p(z|s)$ and $p(s|z)$.
We turn to variational inference techniques for this purpose, and Variational Autoencoders (VAE)~\cite{kingma2014_vae} in particular. 
Fortunately, we can approximate both distributions by training a VAE on samples from $p(s)$ -- the encoder $q_{\psi}$ models $p(z|s)$, whereas the decoder $q_{\phi}$ models $p(s|z)$.
Intuitively, this process determines which latent codes are assigned to each region of the state space, and which states should be visited by each skill. 
The fact that exploration and skill discovery are disentangled enables learning variational posteriors for different $p(z)$ priors without needing to re-learn a new skill-condition policy every time. 
This is an interesting property, as the task of defining the prior over skills is not straightforward.
In contrast, previous methods perform exploration and skill discovery at the same time, so that modifying $p(z)$ inevitably involves exploring the environment from scratch.

\textbf{Skill learning.} The final stage consists in training a policy $\pi_{\theta}(s,z)$ that maximizes the mutual information between states and latent variables. EDL adopts the forward form of the mutual information. The reader is referred to the SM for a detailed explanation of this choice. Since $p(s)$ is fixed, Equation~\ref{eq:forward_mi_bound} can be maximized in a reinforcement learning-styled setup with the reward function
\begin{align}
    r(s,z') = \log q_{\phi}(s|z'), \, z' \sim p(z)  \label{eq:edl_reward}
\end{align}
where $q_{\phi}(s|z)$ is given by the decoder of the VAE trained on the skill discovery stage. 
% Note that an alternative formulation where $z' \sim p(z|s'), s'\sim p(s)$ is also valid, which is equivalent to sampling from the prior when considering a fully-converged VAE.
This final stage can be seen as training a policy that mimics the decoder \textit{within} the MDP, i.e.~a policy that will visit the state that the decoder would generate for each latent code $z$. Note that the reward function is fixed, unlike that in previous methods which continuously changes depending on the behavior of the policy.
% Previous methods continuously perform skill discovery as new states are discovered, thus the policy optimizes an ever-changing reward function.
% The fact that skills are discovered before training the policy results in a reward function that is fixed throughout policy optimization. 
% This might result in more stable training dynamics than with previous methods, as they optimize an ever-changing reward function.

\section{Experiments}
\label{sec:experiments}

% In order to understand the properties and limitations of skill discovery methods, we evaluate them in controlled environments. 
Some previous works have evaluated skill discovery methods on complex environments, such as robotic locomotion~\cite{todorov2012_mujoco} or 3D navigation~\cite{beattie2016_dmlab}, whose complexity renders policy learning difficult. This burden falls on the underlying RL algorithm, which needs to learn a more complicated policy in order to achieve the desired behavior. Note that this does not necessarily make the task of discovering options more difficult. As an example, consider the process of discovering useful locomotion skills. These options will likely require the agent to move in different directions, no matter if it is controlling a simple point mass or a complex humanoid. 

In this work, we take a different approach and consider controlled synthetic environments.
These are fully-continuous 2D mazes where the agent observes its current position and outputs actions that control its location, which is affected by collisions with walls. Varying the maze topology allows for an analysis of skill discovery methods in the face of specific challenges, providing insight on the properties and limitations of these algorithms. 

% Experimental setup
All experiments consider discrete priors over skills. This choice allows for a fair comparison between methods, as those based on the reverse form of the mutual information are not straightforward to combine with continuous priors. We consider the two methods described in Section~\ref{sec:mi_based_methods} as baselines.
% Note that methods based on the forward form of the mutual information, including EDL, do support continuous priors. 
The skill discovery stage in EDL is performed with a VQ-VAE~\cite{vandenoord2017_vqvae} to handle the discrete prior, and the real-valued codes it discovers are used to condition the policy. 
% The reward function in EDL is highly deceptive in some environments, as it relies on the Euclidean distance between states and does not consider the actual connectivity within the MDP. 
We adopt the common distributional assumption for continuous
data where $p(s|z)$ is Gaussian~\cite{kingma2014_vae}, which does not consider the actual connectivity within the MDP and results in reward functions that can become fraught with local optima.
For this reason, we used Sibling Rivalry~\cite{trott2019_sibling} to escape local optima during the skill learning stage in some environments. 
Note that the described implementation is just a possible solution that follows the proposed paradigm, which is not limited to the specific choices made in our experimental setup.

\begin{figure}[ht]
    \centering
    \resizebox{\linewidth}{!}{\includegraphics{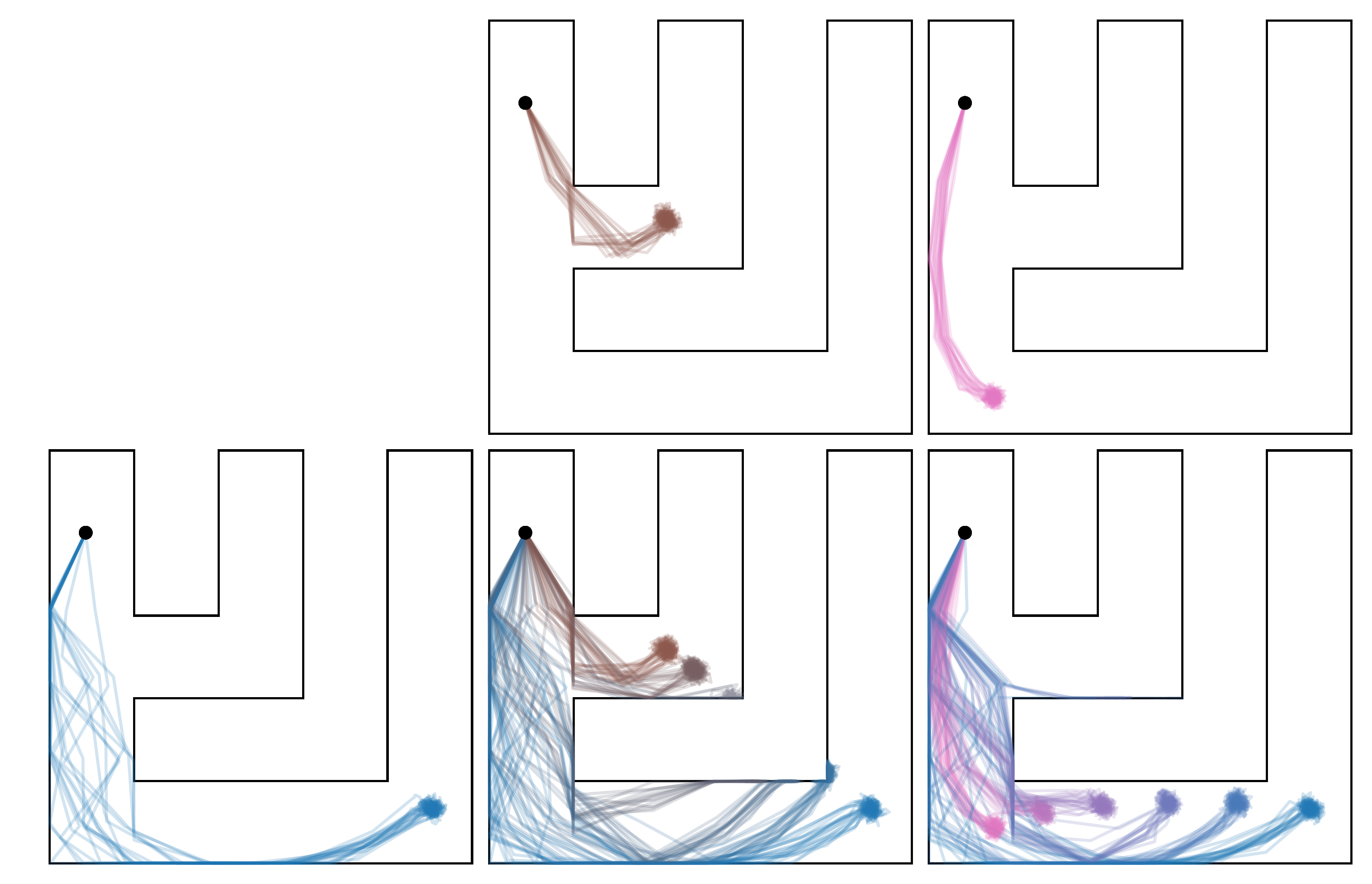}}
    \caption{Interpolating skills learned by EDL. Interpolation is performed at the latent variable level by blending the $z$ vector of two skills. The first row and column show the original skills being interpolated, which were selected randomly from the set of learned options. When plotting interpolated skills, we blend the colors used for the original skills.}
    \label{fig:interpolations}
\end{figure}

% Describe how results are reported and how figures are made
Figures~\ref{fig:motivation}\nobreakdash--\ref{fig:interpolations} visualize multiple rollouts per skill to account for the stochasticity of the policy. The initial state is denoted by a black dot and the color of the rollout denotes the skill upon which it was conditioned, thus figures are best viewed in color. We refer the reader to the SM for a detailed description of the experimental setup and the hyperparameters.

\begin{figure}[ht]
    \centering
    \resizebox{\linewidth}{!}{\includegraphics{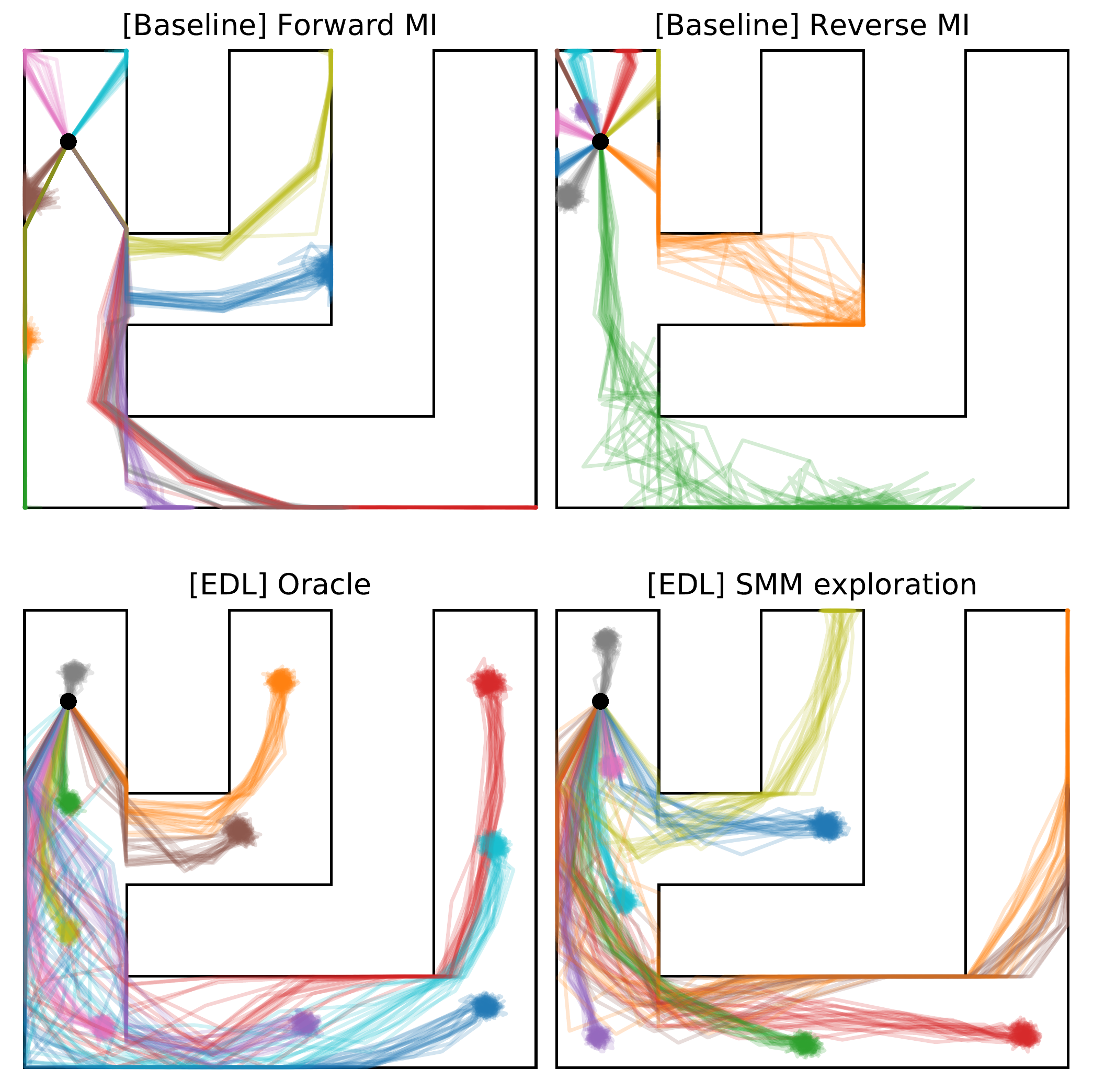}}
    \caption{Impact of replacing the oracle with State Marginal Matching (SMM) in the exploration stage. \textit{Top:} baselines fail at discovering skills that reach the right side of the maze. \textit{Bottom:} EDL discovers skills that are spread across the whole maze, even when replacing the oracle with SMM. We observed that SMM tended to collect more samples near the walls, which explains the slight difference in the discovered options.}
    \label{fig:smm_exploration}
\end{figure}

\textbf{Exploration with SMM.} In the absence of any prior knowledge, we would like to discover skills across the whole state space by defining a uniform distribution over states, $p(s)$. In the controlled environments considered in this work, this can be achieved by sampling states from an oracle. In order to understand the impact of not having access to an oracle, we employ State Marginal Matching (SMM)~\cite{lee2019_smm} with a uniform target distribution to perform the exploration stage in EDL. Evaluation is performed on a simple maze where the forward and reverse baselines already fail to learn state-covering skills, as depicted in Figure~\ref{fig:smm_exploration} (top). In contrast, Figure~\ref{fig:smm_exploration} (bottom) shows how EDL can learn state-covering skills even in the realistic scenario where an oracle is not available\footnote{We succeeded at training skills discovered by EDL without Sibling Rivalry. However, it greatly reduced the number of runs in the grid search that got trapped in local optima. The presented results used Sibling Rivalry to take advantage of this fact and reduce variance in the results.}.

\begin{figure}[ht]
    \centering
    \resizebox{\linewidth}{!}{\includegraphics{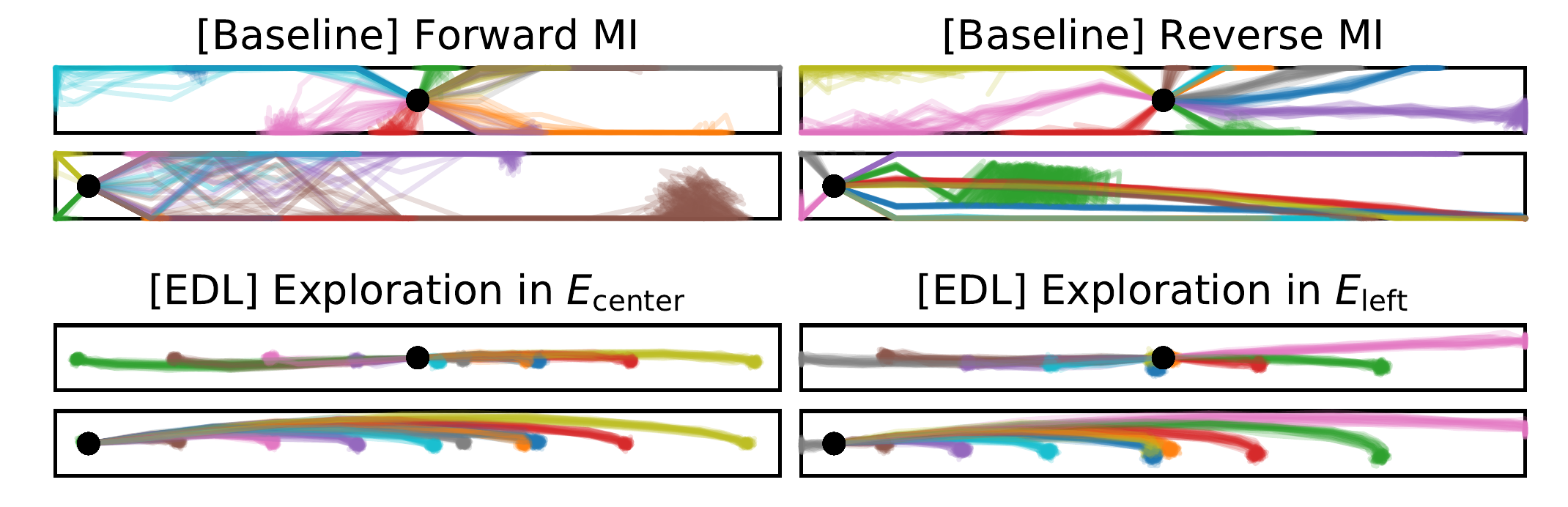}}
    \caption{Impact of the distribution over initial states, $p(s_0)$. \textit{Top:} baselines are very sensitive to $p(s_0)$ and discover very different skills depending on this distribution. 
    \textit{Bottom:} we report two different experiments with EDL. The setup on the left performs exploration and skill discovery in $E_{\text{center}}$ and then learns skills in both $E_{\text{center}}$ and $E_{\text{left}}$. The one on the right performs exploration and skill discovery in $E_{\text{left}}$ instead.
    Options discovered by EDL are very similar in both setups.}
    \label{fig:corridor_varying_s0}
\end{figure}

\textbf{Impact of the initial state.} Baseline methods rely on $\rho_{\pi}(s|z)$ to perform skill discovery, which is initially induced by a random policy. This introduces a strong dependence on the distribution over initial states, $p(s_0)$. Changes to $p(s_0)$ might make some behaviors harder to learn, e.g.~reaching a certain position becomes more difficult the further an agent spawns from it. As long as all options are still achievable, a change in $p(s_0)$ should have little impact on what options are deemed important. We evaluate this phenomenon on two corridor-shaped mazes, which have the same topology but differ in the position of the initial state. 
We will refer to these environments as $E_{\text{center}}$ and $E_{\text{left}}$, in which the agent spawns in the center and the left section of the corridor, respectively. 
Figure~\ref{fig:corridor_varying_s0} (top) shows how the baselines discover completely different skills depending on $p(s_0)$. 
When replicating this experiment using EDL with SMM exploration, we get two different setups. Figure~\ref{fig:corridor_varying_s0} (bottom left) shows the result of performing exploration and skill discovery in $E_{\text{center}}$ and then learning skills in both $E_{\text{center}}$ and $E_{\text{left}}$. Figure~\ref{fig:corridor_varying_s0} (bottom right) depicts the impact of performing exploration and skill discovery in $E_{\text{left}}$ instead.
Skills learned in both setups are very similar, with differences coming from the slightly different distribution over states collected by SMM.

\begin{figure}[ht]
    \centering
    \resizebox{\linewidth}{!}{\includegraphics{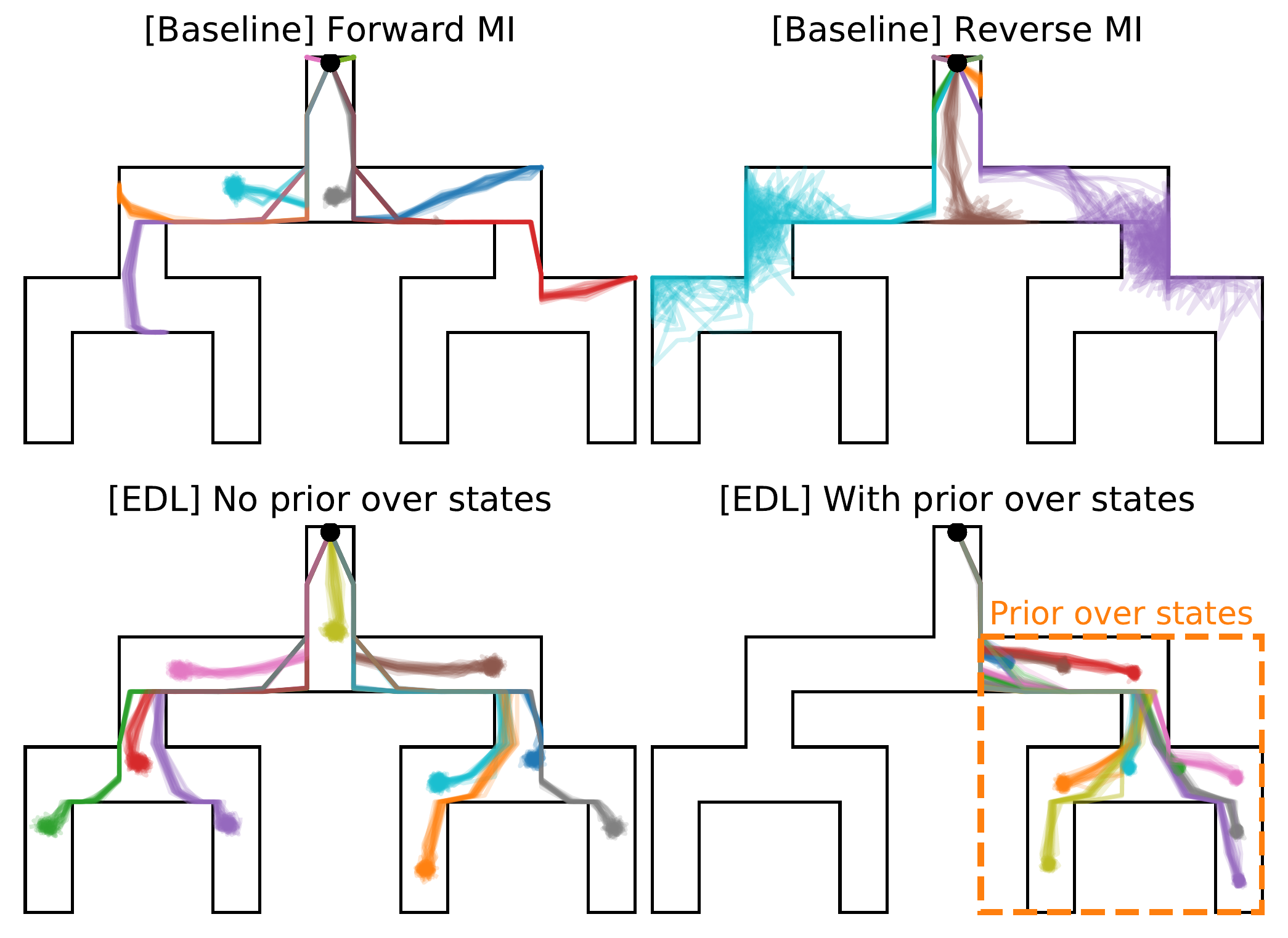}}
    \caption{Incorporating priors over skills, where we are interested in learning skills on the right side of the maze. \textit{Top:} this type of prior cannot be incorporated into baseline methods, whose discovered options are agnostic to it. \textit{Bottom:} in the absence of a prior, EDL learns options across the whole state space. When incorporating the prior, the agent devotes all its capacity to learning skills within the region of interest.}
    \label{fig:prior_on_p_s}
\end{figure}

\textbf{Encouraging specific behaviors.}
% So far we considered the scenario where the agent needs to discover skills without any prior specification of what makes an option useful.
In many settings, the user has some knowledge about
which areas of state space will be most relevant for downstream tasks.
% the downstream tasks for which these options will be used.
Existing methods can leverage prior knowledge by maximizing $I(f(S);Z)$ instead of $I(S;Z)$, where $f(S)$ is a function of the states. For instance, this function can compute the center of mass of a robot in order to encourage the discovery of locomotion skills~\cite{eysenbach2019_diayn}. However, this method fails at incorporating more complex priors, such as encouraging the agent to only learn locomotion skills that move in specific directions. EDL offers more flexibility for leveraging priors through the definition of $p(s)$, e.g.~by drawing samples from a dataset of human play~\cite{guss2019_minerl}. We simulate this scenario by performing skill discovery with an oracle that samples states uniformly from a subset of the state space. Figure~\ref{fig:prior_on_p_s} reports results in a tree-shaped maze, where we introduce the prior that skills should visit the right side of the maze only. EDL effectively incorporates this prior, and learns state-covering skills in its absence. 

\textbf{Impact of bottleneck states.} The maze with bottleneck states from Figure~\ref{fig:motivation}, where baseline approaches fail to explore a large extent of the state space, is a challenging environment where the limitations of EDL can be evaluated. We were unable to explore this type of maze effectively with SMM. Given that SMM relies on a curiosity-like bonus~\cite{lee2019_smm}, we attribute this failure to well-known issues of these methods such as derailment and detachment~\cite{ecoffet2019_goexplore}. Note that these problems are related to the sub-optimality of the density estimation method and RL solver, as shown by the bounds derived by \citet{hazan2019_maxent}. 
% In light of these issues, \citet{ecoffet2019_goexplore} propose a method that exploits determinism during training in simulated environments to improve exploration. Instead of exploiting environment-specific properties, we rely on an oracle in order to simulate perfect exploration and evaluate the rest of stages in EDL.
In light of this, we rely on an oracle to simulate perfect exploration and evaluate the skill discovery/learning stages of EDL. On this maze, the reward functions that EDL introduces create deceptive local optima in which policies tend to get stuck (visualizations are reported in the SM). Sibling Rivalry proved crucial to avoid these failures, and allowed the policy to learn the skills depicted in Figure~\ref{fig:bottleneck_maze}~(bottom right). These observations suggest that the main bottlenecks for the proposed approach to skill discovery are maximum entropy exploration and avoiding local optima when learning to maximize the EDL reward (Equation \ref{eq:edl_reward}). Given that EDL decouples the process in three stages, advances in these fields are straightforward to incorporate and will boost the performance of this type of option discovery method.

\textbf{Interpolating between skills.} The skill discovery stage in EDL with a categorical prior $p(z)$ can be seen as the process of learning a discrete number of goals, together with an embedded representation for each of them. 
In the experimental setup presented in this work, each embedded representation corresponds to one of the continuous vectors in the VQ-VAE's codebook, $z_i$, whereas each goal state is given by $g_i = \text{argmax}_s q_{\phi}(s|z_i)$. The idea of goal embeddings was introduced as part the Universal Value Function Approximators (UVFA) framework~\cite{schaul2015_uvf}. UVFAs can generalize to unseen goals, and here we explore how the policies learned by EDL generalize to unseen latents $z$ -- where we construct new latents by interpolating the ones discovered by EDL. The results in Figure~\ref{fig:interpolations} suggest that the policy learns to generalize, with interpolated skills reaching states that come from the interpolation in Euclidean space of the goals of the original skills. Additional visualizations are included in the SM.

\section{Related work}

\textbf{Option discovery.} Temporally-extended high-level primitives, also known as options, are an important resource in the RL toolbox~\cite{parr1998_hierarchies,sutton1999_between,precup2001_temporal}. The process of defining options involves task-specific knowledge, which might be difficult to acquire and has motivated research towards methods that automatically discover such options. These include learning options while solving the desired task~\cite{bacon2017_option}, leveraging demonstrations~\cite{fox2017_multilevel}, training goal-oriented low-level policies~\cite{nachum2018_data-efficient-hrl}, and meta-learning primitives from a distribution of related tasks~\cite{frans2018_meta}. 
Skills discovered by information-theoretic methods 
% such as the ones considered in this work 
have also been used as primitives for Hierarchical RL~\cite{florensa2017_snn,eysenbach2019_diayn,sharma2019_dynamics}.
% Methods considered in this work discover skills in an unsupervised fashion, which can be used as primitives for Hierarchical RL~\cite{florensa2017_snn,eysenbach2019_diayn}.

\textbf{Intrinsic rewards.} Agents need to encounter a reward before they can start learning, but this process might become highly inefficient in sparse reward setups when relying on standard exploration techniques~\cite{osband2016_generalization}. This issue can be alleviated by introducing intrinsic rewards, i.e.~denser reward signals that can be automatically computed. These rewards are generally task-agnostic and might come from state visitation pseudo-counts~\cite{bellemare2016_unifying,tang2017_exploration}, unsupervised control tasks~\cite{jaderberg2017_unreal}, learning to predict environment dynamics~\cite{houthooft2016_vime,pathak2017_curiosity,burda2018_rnd}, self-imitation~\cite{oh2018_sil}, and self-play~\cite{sukhbaatar2017_intrinsic,liu2018_cer}.

\textbf{Novelty Search.} Discovering a set of diverse and task-agnostic behaviors in the absence of a fitness function has been explored in the evolutionary computation community~\cite{lehman2011_abandoning,lehman2011_novelty}. Quality Diversity algorithms aim at combining the best of both worlds by optimizing task-specific fitness functions while encouraging diverse behaviors in a population of agents~\cite{cully2015_robots,mouret2015_illuminating,pugh2016_quality}. These methods rely on a behavior characterization function, which is tasked with summarizing the behavior of an agent into a vector representation. There have been efforts towards learning such functions~\cite{meyerson2016_learning-bc}, but it is still a common practice for practitioners to design a different function for each task~\cite{conti2017_novelty}.

\textbf{Goal-oriented RL.} The standard RL framework can be extended to consider policies and reward functions that are conditioned on some goal $g \in \mathcal{G}$~\cite{schaul2015_uvf}. Given a known distribution over goals $p(g)$ that the agent should achieve, this setup allows for efficient training techniques involving experience relabeling~\cite{andrychowicz2017_her} and reward shaping~\cite{trott2019_sibling}. Defining such distribution requires expert domain knowledge, an assumption that is not always fulfilled. As a result, methods that can learn to reach \textit{any} given state have garnered research interest~\cite{warde2019_discern,pong2019_skewfit}. These approaches can be seen as skill discovery algorithms where $\mathcal{Z} = \mathcal{S}$, i.e.~where each goal state defines a different skill. This raises the question of whether methods that can reach any state are superior to those learning a handful of skills. We argue that the latter offer important benefits in terms of exploration when used by a meta-controller to solve downstream tasks. 
When $p(z)$ is a simple distribution, the meta-controller benefits from a reduced search space, which is one of the motivations behind building hierarchies and options~\cite{precup2001_temporal}. 
% The search space is reduced when $p(z)$ is simple, which is one of the motivations behind building hierarchies and options~\cite{precup2001_temporal}. 
On the other hand, exploring with state-reaching policies involves a search space of size $|\mathcal{S}|$. This figure will quickly increase as the complexity of the environment grows, making exploration inefficient. 
Moreover, this setup assumes that the meta-controller is to be able to sample from $\mathcal{S}$ in order to emit goals for the goal-conditioned policy.

\section{Discussion}

% Summary of findings
We provide theoretical and empirical evidence that poor state space coverage is a predominant failure mode of existing skill discovery methods. The information-theoretic objective requires access to unknown distributions, which these methods approximate with those induced by the policy. These approximations lead to pathological training dynamics where the agent obtains larger rewards by visiting already discovered states rather than exploring the environment. 
We propose \textit{Explore, Discover and Learn} (EDL), a novel option discovery approach that leverages a fixed distribution over states and variational inference techniques to break the dependency on the distributions induced by the policy. 
Importantly, this alternative approach optimizes the same objective derived from information theory used in previous methods.
EDL succeeds at discovering state-covering skills in environments where previous methods failed. It offers additional advantages, such as being more robust to changes in the distribution of the initial state and enabling the user to incorporate priors over which behaviors are considered useful. Our experiments suggest that EDL discovers a meaningful latent space for skills even when tasked with learning a discrete set of options, whose latent codes can be combined in order to produce a richer set of behaviors. 

% EDL is a new paradigm which is not limited to the proposed implementation
The proposed EDL paradigm is not limited to the implementation considered in this work. Each of the three stages of the method poses its own challenges, but can benefit from advances in their respective research directions as well. This modular design allows us to incorporate to our implementation recent advances such as exploration with State Marginal Matching~\cite{lee2019_smm}, vector quantization techniques to impose discrete priors in VAEs~\cite{vandenoord2017_vqvae}, and relabeling techniques to optimize deceptive reward functions~\cite{trott2019_sibling}. Future breakthroughs in these directions could contribute towards scaling up skill discovery methods to richer environments, potentially leading to the emergence of complex behaviors~\cite{jaderberg2019_ftw}.

% List some ideas for future research that can be incorporated into EDL
There are several research directions to be explored in future work. Improvements in pure exploration methods would make EDL applicable to a broader range of environments. Despite the existence of strong theoretical results~\cite{hazan2019_maxent}, these approaches involve the optimization of reward functions that are challenging for current algorithms~\cite{ecoffet2019_goexplore}. In our experiments, we adopted the common distributional assumption for continuous data where $p(s|z)$ is Gaussian~\cite{kingma2014_vae}. This assumption was responsible for the deceptive reward functions discovered in our experiments, and might be detrimental in some other environments. This motivates research towards discovering embedding spaces for states where distances are related to the closeness of states within the MDP~\cite{florensa2019_self}, and learning reward functions that reflect similarity in controllable aspects of the environment~\cite{warde2019_discern}. Finally, leveraging information-theoretic methods to perform unsupervised task discovery in the meta-RL framework~\cite{jabri2019_unsupervised} is another interesting direction for future research.

\section*{Acknowledgements}

%%% FUNDING
%% Xavi's funding (GPI/TSC): 
%%       industrial doctorate 2017 DI 011 - Generalitat de Catalunya
%%       contract TEC2016-75976-R - Spanish Ministry of Science and Innovation and the European Regional Development Fund (ERDF)
%% Jordi's funding (DAC/BSC): 
%%       contract TIN2015-65316-P - Spanish Ministry of Science and Innovation
%%		 BSC-CNS Severo Ochoa program SEV-2015-0493
%%		 contract 2017-SGR-1414 - Generalitat de Catalunya
%% Victor: La Caixa Scholarship

This work was partially supported by the Spanish Ministry of Science and Innovation and the European Regional Development Fund under contracts TEC2016-75976-R and TIN2015-65316-P, by the BSC-CNS Severo Ochoa program SEV-2015-0493, and by Generalitat de Catalunya under contracts 2017-SGR-1414 and 2017-DI-011.
V{\'\i}ctor Campos was supported by Obra Social ``la Caixa'' through La Caixa-Severo Ochoa International Doctoral Fellowship program.
  %% Removed from the submission

\bibliography{references}
\bibliographystyle{icml2020}

%% ICML does not accept anything after the references
%% We can use pdftk to extract the SM while keeping the references
%% Sample command to extract pages 12-end (references end at page 11):
%%   pdftk complete_paper.pdf cat 12-end output supplementary.pdf

%%%%%%%%%%%%%%%% APPENDIX %%%%%%%%%%%%%%%%

\clearpage
\appendix
\section{Theoretical analysis of existing methods}
\label{sec:theoretical_rew_analysis}

\begin{figure}[ht]
    \centering
    
    \BlankLine
    % $\rho_{\pi}(s|z)$ \\
    \begin{tabularx}{\linewidth}{YY}
       $\rho_{\pi}(s|z=z_0)$  & $\rho_{\pi}(s|z=z_1)$
    \end{tabularx}
    \resizebox{.48\linewidth}{!}{\includegraphics{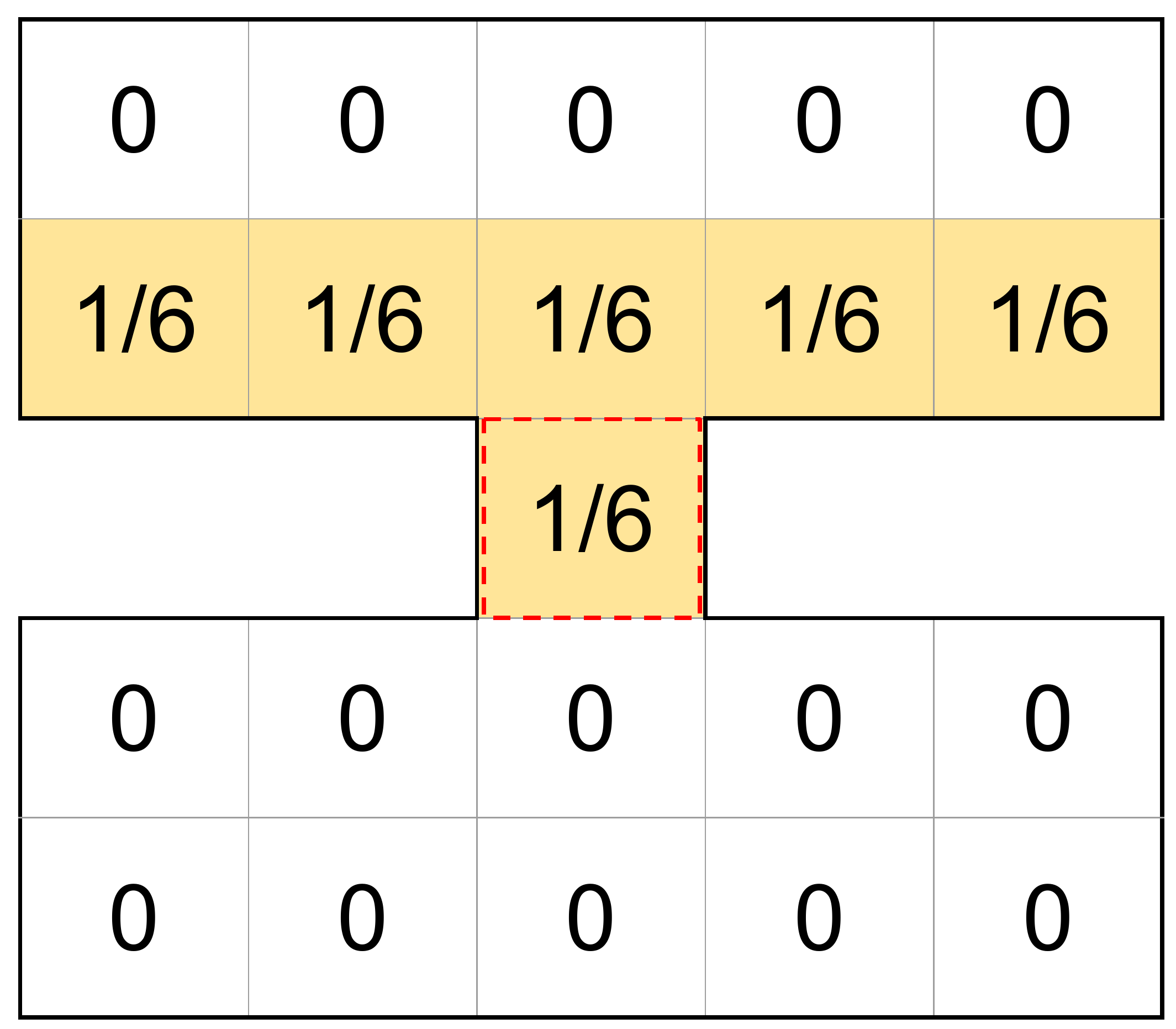}}
    \resizebox{.48\linewidth}{!}{\includegraphics{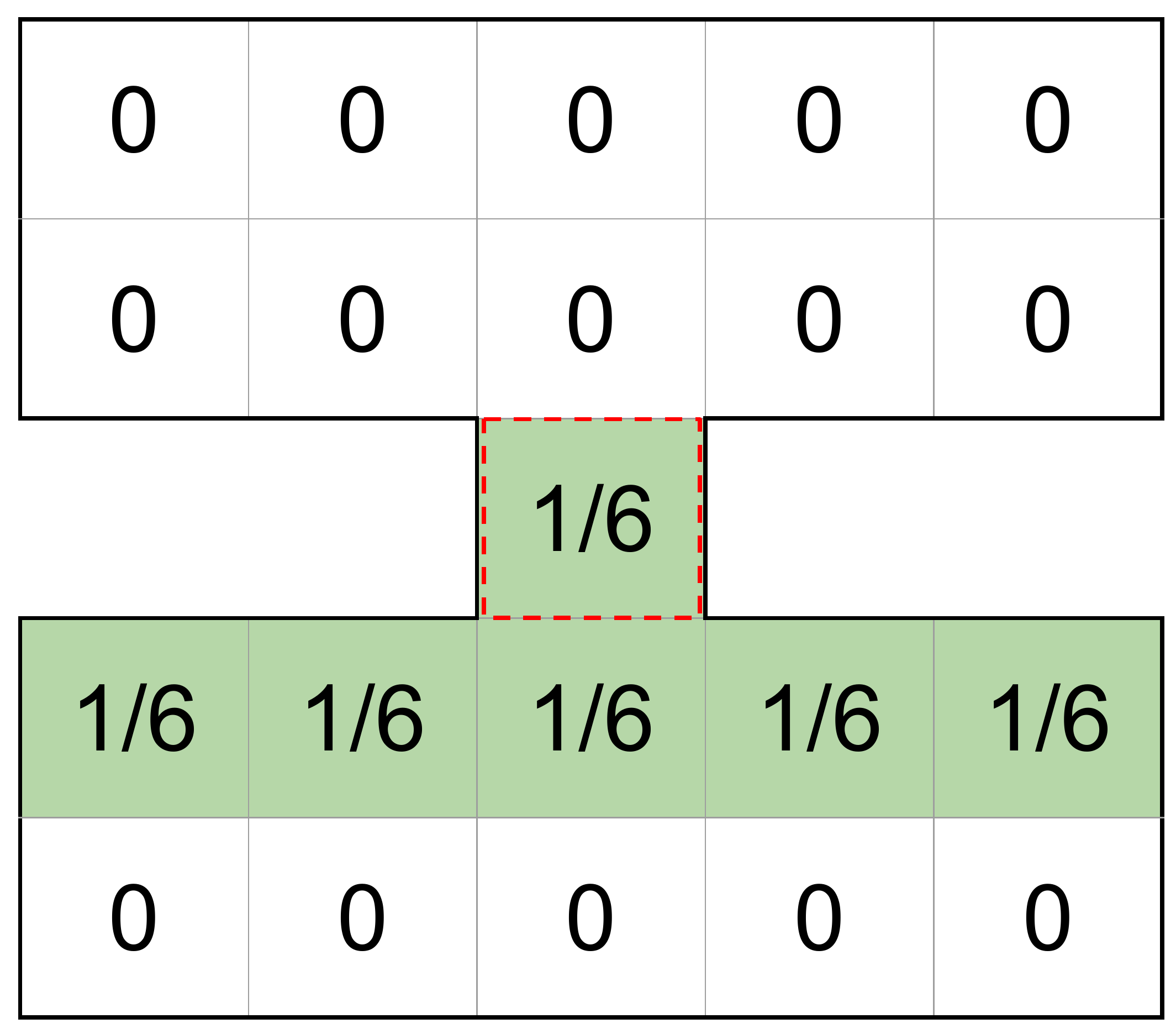}}
    
    \BlankLine
    % $r(s,z)$ \\
    \begin{tabularx}{\linewidth}{YY}
       $r(s,z=z_0)$  & $r(s,z=z_1)$
    \end{tabularx}
    \resizebox{.48\linewidth}{!}{\includegraphics{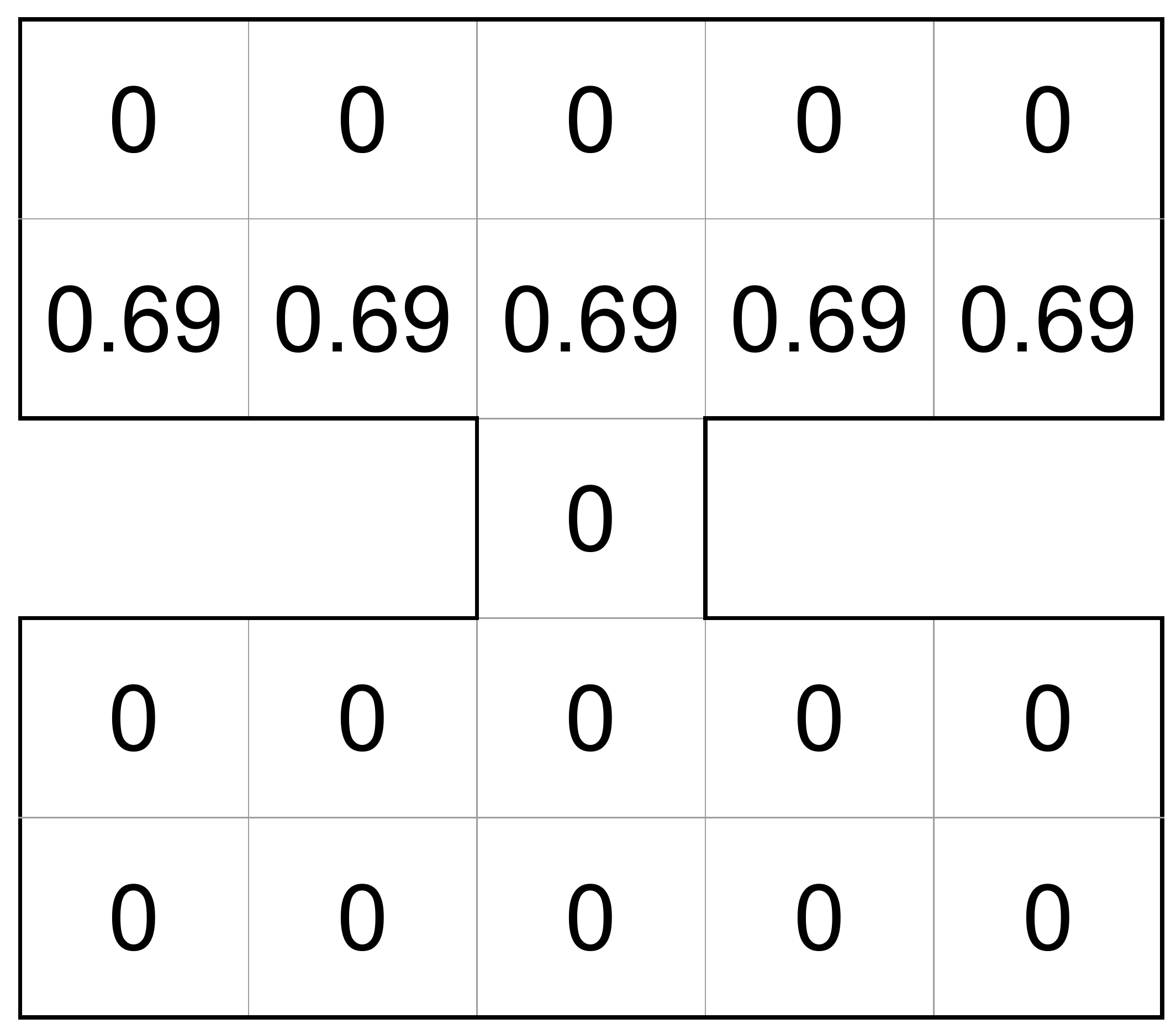}}
    \resizebox{.48\linewidth}{!}{\includegraphics{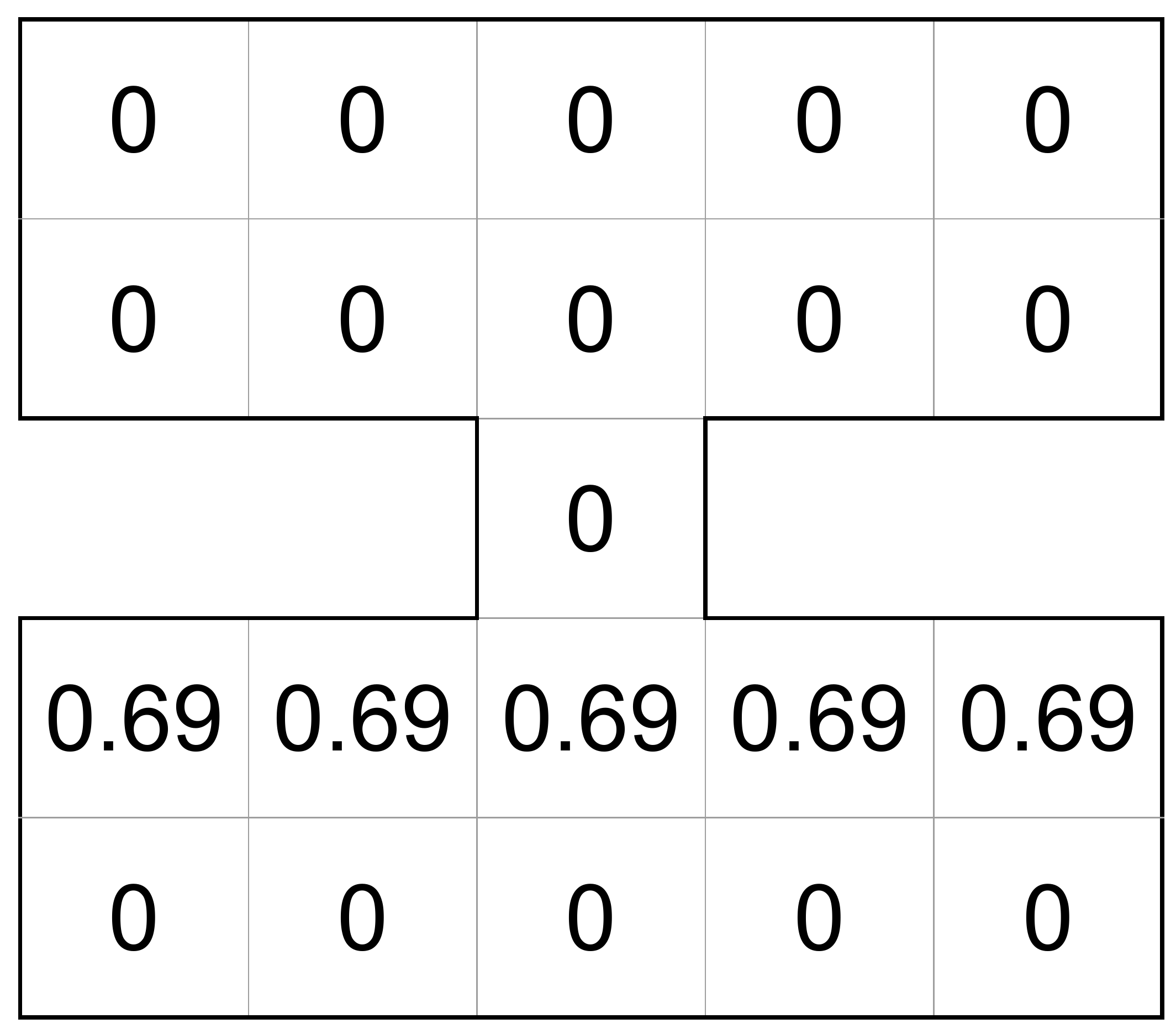}}
    
    \caption{Analysis of the reward landscape on a toy gridworld with two skills, assuming perfect density estimation. Under this assumption, both forms of the mutual information generate the same reward landscape. Each column depicts a different skill, and all rollouts always start from the central tile which is highlighted in red. Skills are rewarded for visiting known states where they are maximally distinguishable, but receive no reward for visiting novel states. }
    \label{fig:gridworld}
\end{figure}

This section provides insight for why existing methods do not encourage the discovery state-covering skills from a theoretical lens. This is achieved by analyzing the reward function of these methods, and studying its asymptotic behavior for known and novel states.
Our main result shows that the agent receives larger rewards for visiting known states than discovering new ones.
The following subsections contain the derivation of this result, and Figure~\ref{fig:gridworld} provides a numerical example on a gridworld environment.

\subsection{Reverse form of the mutual information}

The objective for these methods is
\begin{align}
    I(S;Z) &= \mathbb{E}_{s,z \sim p(s,z)} [\log p(z|s)] - \mathbb{E}_{z \sim p(z)}[\log p(z)] \\
    &\approx \mathbb{E}_{s,z \sim p(s,z)} [\log \rho_{\pi}(z|s)] - \mathbb{E}_{z \sim p(z)}[\log p(z)]
\end{align}
where the unknown posterior $p(z|s)$ is approximated by the distribution induced by the policy, $\rho_{\pi}(z|s)$. This distribution is estimated with a model $q_{\theta}(z|s)$ trained via maximum likelihood on $(s,z)$-tuples collected by deploying the policy in the environment.
% In practice, the posterior $\rho_{\pi}(z|s)$ induced by the policy is approximated by a model $q_{\theta}(z|s)$ trained via maximum likelihood. 
For this analysis, however, we will assume access to a perfect estimate of $\rho_{\pi}(z|s)$. When considering the discovery of $N$ discrete skills under a uniform prior, the reward in Equation~\ref{eq:reverse_mi_rew} becomes
\begin{align}
    r(s,z') &= \log \rho_{\pi}(z'|s) - \log p(z') \\
    &= \log \rho_{\pi}(z'|s) + \log N
\end{align}
where $z' \sim p(z)$. We will assume that $\sum_{i=1}^{N}\rho_{\pi}(z_i|s)=1$ in our analysis.

\textbf{Maximum reward for known states.} The reward function encourages policies to discover skills that visit disjoint regions of the state space where $\rho_{\pi}(z'|s) \to 1$:
%Therefore, states that are visited by a single skill provide the maximum reward, $r = \log N$.
\begin{align}
    r_{\text{max}} %&= \lim_{\rho_{\pi}(z'|s) \to 1} \log \rho_{\pi}(z'|s) + \log N \\
    =\log 1 + \log N = \log N
\end{align}

\textbf{Reward for previously unseen states.} Note that $\rho_{\pi}(z|s)$ is not defined for unseen states, and we will assume a uniform prior over skills in this undefined scenario, $\rho_{\pi}(z|s) = 1/N, \forall z$:
\begin{align}
    r_{\text{new}} = \log \frac{1}{N} + \log N = 0
\end{align}

Alternatively, one could add a \textit{background} class to the model in order to assign null probability to unseen states~\cite{capdevila2018_mining}. This differs from the setup in previous works, reason why it was considered in the analysis. However, note that the agent gets a larger penalization for visiting new states in this scenario:
\begin{align}
    r_{\text{new}}^{'} = \lim_{\rho_{\pi}(z'|s) \to 0} \log \rho_{\pi}(z'|s) + \log N = - \infty
\end{align}

These observations explain why the learned skills provide a poor coverage of the state space.

\subsection{Forward form of the mutual information}

The objective for these methods is
\begin{align}
    I(S;Z) &= \mathbb{E}_{s,z \sim p(s,z)} [\log p(s|z)] - \mathbb{E}_{s \sim p(s)}[\log p(s)] \\
    &= \mathbb{E}_{s,z \sim p(s,z)} [\log \rho_{\pi}(s|z)] - \mathbb{E}_{s \sim \rho_{\pi}(s)}[\log \rho_{\pi}(s)]
\end{align}
where the unknown distributions $p(s|z)$ and $p(s)$ are approximated using the stationary state-distribution, $p(s|z) \approx \rho_{\pi}(s|z)$ and $p(s) \approx \rho_{\pi}(s) = \mathbb{E}_z \left[ \rho_{\pi}(s|z) \right]$. The stationary state-distribution is estimated with a model $q_{\theta}(s|z)$ trained via maximum likelihood on $(s,z)$-tuples collected by deploying the policy in the environment.
% In practice, the stationary state-distribution is approximated by a model $q_{\theta}(s|z)$ trained via maximum likelihood. 
For this analysis, however, we will assume access to a perfect estimate of $\rho_{\pi}(s|z)$. When considering the discovery of $N$ discrete skills, the reward in Equation~\ref{eq:forward_mi_rew} can be expanded as follows:
\begin{align}
    r(s,z') &= \log \rho_{\pi}(s|z') - \log \frac{1}{N} \sum_{\forall z_i} \rho_{\pi}(s|z_i) \\
    &= \log \frac{\rho_{\pi}(s|z')}{\sum_{\forall z_i} \rho_{\pi}(s|z_i)} + \log N \label{eq:rew_forward_mi_softmax} \\
    % &\xrightarrow{\epsilon \to 0} \log \frac{1}{1 + \sum_{\forall z_i \neq z'} \frac{\rho_{\pi}(s|z_i) + \epsilon}{\rho_{\pi}(s|z') + \epsilon}} + \log N \label{eq:rew_forward_mi_large_frac}
    &= \lim_{\epsilon \to 0} \log \frac{1}{1 + \sum_{\forall z_i \neq z'} \frac{\rho_{\pi}(s|z_i) + \epsilon}{\rho_{\pi}(s|z') + \epsilon}} + \log N \label{eq:rew_forward_mi_large_frac}
\end{align}
where $z', z_i \sim p(z)$ and we added $\epsilon \to 0$ in the last step to prevent division by 0. 

\textbf{Maximum reward for known states.} As observed by \citet{sharma2019_dynamics}, this reward function encourages skills to be predictable (i.e.~$\rho_{\pi}(s|z') \to 1$) and diverse (i.e.~$\rho_{\pi}(s|z_i) \to 0, \forall z_i \neq z'$):
\begin{align}
    r_{\text{max}} = \log 1 + \log N = \log N
\end{align}

\textbf{Reward for previously unseen states.} In novel states, $\rho_{\pi}(s|z_i) \to 0, \forall z_i$: 
\begin{align}
    r_{\text{max}} &= \lim_{\epsilon \to 0} \log \frac{1}{1 + \sum_{\forall z_i \neq z'} \frac{\epsilon}{\epsilon}} + \log N \\
    &= \log \frac{1}{1 + (N - 1)} + \log N \\
    &= \log \frac{1}{N} + \log N \\
    &= 0
\end{align}

This result shows that visiting known states instead of exploring unseen ones provides larger rewards to the agent, producing options that provide a poor coverage of the state space.

\section{Choice of mutual information's form}
\label{sec:choice_of_mi_form}

The main novelty of EDL is an alternative for modelling the unknown distributions, which in principle could work with either form of the mutual information. For the sake of comparison with previous works, all experiments consider discrete skills. This was achieved through a categorical posterior $p(z|s)$ that was approximated with a VQ-VAE~\cite{vandenoord2017_vqvae}. The encoder of the VQ-VAE takes an input $x$, produces output $z_e(x)$, and maps it to the closest element in the codebook, $e \in \mathbb{R}^{K \times D}$. The posterior categorical distribution $q(z|x)$ probabilities are defined as one-hot as follows:
\begin{equation}
    q(z=k|x) = 
    \begin{cases}
        1 & \text{for } k = \text{argmin}_j ||z_e(x) - e_j||_2  \\
        0 & \text{otherwise}
  \end{cases}
\end{equation}

One could consider the reverse form of the mutual information and train the policy with a reward function as follows:
\begin{equation}
    r(s,z) = q(z|s)
\end{equation}
where we assumed a uniform prior over $z$ and removed the constant $\log p(z)$ term from the reward. 

We can foresee two issues with this reward function. It is sparse, i.e.~many states provide no reward at all, which might hinder training unless proper exploration strategies are used~\cite{ecoffet2019_goexplore,trott2019_sibling}. A similar behavior was observed in existing methods using the reverse form of the mutual information (c.f.~Figure~\ref{fig:bottleneck_maze-reverse_mi_reward_landscape}). Moreover, the fact that many states produce a maximum reward of 1 might lead to unpredictable skills when paired with an entropy bonus. Such unpredictability might not be desirable when training a metra-controller to solve a downstream task by combining the learned skills~\cite{sharma2019_dynamics}.

\section{Implementation Details}
\label{sec:implementation_details}

\textbf{Environments.} The maze environments are adapted from the open-source implementation\footnote{\url{https://github.com/salesforce/sibling-rivalry}} by \citet{trott2019_sibling}. The agent does not observe the walls, whose location needs to be inferred from experience and makes exploration difficult. The initial state for each episode is sampled from a $1 \times 1$ tile. See Table~\ref{tab:env_details} for details about the environments and the topology of each maze.

\begin{table}[ht]
    \centering
    \begin{tabularx}{\linewidth}{lL}
         \toprule
         \textbf{Parameter} & \textbf{Value} \\
         \midrule
         State space & $\mathcal{S} \in \mathbb{R}^2$  \\
         Action space & $\mathcal{A} \in [-0.95, 0.95]^2$  \\
         Episode length & $50$  \\
         \midrule
         Size: Bottleneck maze (Figure~\ref{fig:motivation}) & $10 \times 10$  \\
         Size: Square maze (Figure~\ref{fig:smm_exploration}) & $5 \times 5$  \\
         Size: Corridor maze (Figure~\ref{fig:corridor_varying_s0}) & $1 \times 12$  \\
         Size: Tree maze (Figure~\ref{fig:prior_on_p_s}) & $7 \times 7$  \\
         \bottomrule
    \end{tabularx}
    \caption{Environment details.}
    \label{tab:env_details}
\end{table}

\textbf{RL Agents.} Policy networks emit the parameters of a Beta distribution~\cite{chou2017_beta}, which are then shifted and scaled to match the task action range. Entropy regularization is employed to prevent convergence to deterministic behaviors early in training. We use a categorical distribution with uniform probabilities for the skill prior $p(z)$. 
Agents are trained with PPO~\cite{schulman2017_ppo} and the Adam optimizer~\cite{kingma2014_adam}. Hyperparameters are tuned for each method independently using a grid search. See Table~\ref{tab:hyperparameters} for details.

\begin{table}[ht]
    \centering
    \begin{tabularx}{\linewidth}{lL}
         \toprule
         \textbf{Hyperparameter} & \textbf{Value} \\
         \midrule
         Discount factor & $0.99$  \\
         $\lambda_{\text{GAE}}$ & $0.98$  \\
         $\lambda_{\text{entropy}}$ & $\left\{ 0.001, 0.005, 0.01, 0.025 \right\}$  \\
         $\epsilon_{\text{SiblingRivalry}}$ & $\left\{ 2.5, 5.0, 7.5 \right\}$  \\
         \midrule
         Optimizer & Adam  \\
         Learning rate & $\left\{ 0.0003, 0.001 \right\}$  \\
         Learning rate schedule & Constant \\
         \midrule
         Advantage normalization & Yes \\
         Input normalization & \{Yes, No\}  \\
         \midrule
         Hidden layers & $2$  \\
         Units per layer & $128$  \\
         Non-linearity & ReLU  \\
         \midrule
         Horizon & 2500 \\
         Batch size & 250 \\
         Number of epochs & 4 \\
         \bottomrule
    \end{tabularx}
    \caption{Hyperparameters used in the experiments. Values between brackets were used in the grid search, and tuned independently for each method.}
    \label{tab:hyperparameters}
\end{table}

\textbf{Exploration.} When relying on State Marginal Matching (SMM)~\cite{lee2019_smm} for exploration, we implement the version that considers a mixture of policies with a uniform target distribution $p^*(s)$. The density model $q(s)$ is approximated with a VAE. We use states in the replay buffer as a non-parametric approach to sampling from the desired $p(s)$~\cite{warde2019_discern}. Sampling states from the replay buffer is similar to a uniform Historical Averaging strategy. This worked well in our experiments, but exponential sampling strategies might be needed in other environments to avoid oversampling states collected by the initially random policies~\cite{hazan2019_maxent}. Our implementation follows the open-source code released by the authors\footnote{\url{https://github.com/RLAgent/state-marginal-matching}}, which relies on SAC for policy optimization. Hyperparameters are tuned for each environment independently using a grid search. See Table~\ref{tab:hyperparameters_smm} for details.

\begin{table}[ht]
    \centering
    \begin{tabularx}{\linewidth}{lL}
         \toprule
         \textbf{Hyperparameter} & \textbf{Value} \\
         \midrule
         Discount factor & $0.99$  \\
         Target smoothing coefficient & $0.005$ \\
         Target update interval & $1$ \\
         $\alpha_{\text{entropy}}$ & $\left\{ 0.1, 1, 10 \right\}$  \\
         $\beta_{\text{VAE}}$ & $\left\{ 0.01, 0.1, 1 \right\}$  \\
         \midrule
         Optimizer & Adam  \\
         Policy: Learning rate & $0.001$  \\
         SMM discriminator: Learning rate & $0.001$  \\
         VAE: Learning rate & $0.01$  \\
         Learning rate schedule & Constant \\
         \midrule
         Policies in the mixture & 4 \\
         Input normalization & No  \\
         \midrule
         Policy: Hidden layers & $2$  \\
         SMM discriminator: Hidden layers & $2$  \\
         VAE encoder: Hidden layers & $2$  \\
         VAE decoder: Hidden layers & $2$  \\
         Units per layer & $128$  \\
         Non-linearity & ReLU  \\
         \midrule
         Gradient steps & 1 \\
         Batch size & 128 \\
         Replay buffer size & 50k \\
         \bottomrule
    \end{tabularx}
    \caption{Hyperparameters used for exploration using SMM. Values between brackets were used in the grid search, and tuned independently for each environment. Training ends once the buffer is full.}
    \label{tab:hyperparameters_smm}
\end{table}

\textbf{Skill discovery.} The skill discovery stage in the proposed method is done with a VQ-VAE~\cite{vandenoord2017_vqvae}, which allows learning discrete latents. We implement the version that relies on a commitment loss to learn the dictionary. The size of the codebook is set to the number of desired skills. Hyperparameters are tuned for each environment and exploration method independently using a grid search. See Table~\ref{tab:hyperparameters_vqvae} for details. 

\begin{table}[ht]
    \centering
    \begin{tabularx}{\linewidth}{lL}
         \toprule
         \textbf{Hyperparameter} & \textbf{Value} \\
         \midrule
         Code size & 16 \\
         $\beta_{\text{commitment}}$ & $\left\{ 0.25, 0.5, 0.75, 1.0, 1.25 \right\}$ \\
         \midrule
         Optimizer & Adam  \\
         Learning rate & $0.0002$  \\
         Learning rate schedule & Constant \\
         Batch size & 256 \\
         \midrule
         Number of samples & 4096 \\
         Input normalization & Yes  \\
         \midrule
         Encoder: Hidden layers & $2$  \\
         Decoder: Hidden layers & $2$  \\
         Units per layer & $128$  \\
         Non-linearity & ReLU  \\
         \bottomrule
    \end{tabularx}
    \caption{Hyperparameters used for training the VQ-VAE in the skill discovery stage. Values between brackets were used in the grid search, and tuned independently for each environment and exploration method.}
    \label{tab:hyperparameters_vqvae}
\end{table}

\section{Figure details}

All experiments in the paper consider agents that learn 10 skills. This value was selected to provide a good balance between learning a variety of behaviors and ease of visualization. Given the stochastic nature of the learned policies, we report 20 rollouts per skill. When visualizing states visited by a random policy, we collect 100 rollouts with each (untrained) skill. Trajectories from these skills highly overlap with each other, so we use a single color for all of them to reduce clutter.

\section{Additional visualizations}

\begin{figure}[ht]
    \centering
    \resizebox{.48\linewidth}{!}{\includegraphics{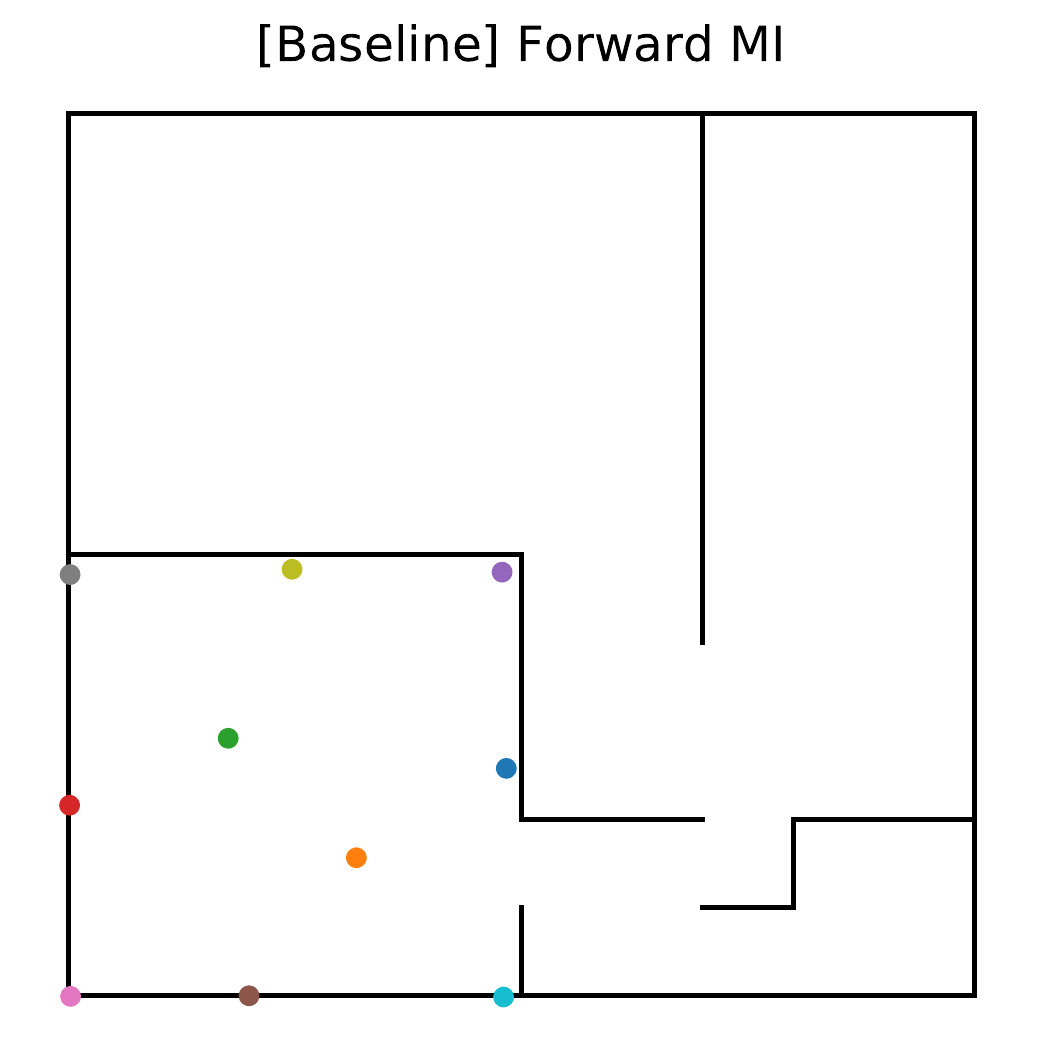}}
    \resizebox{.48\linewidth}{!}{\includegraphics{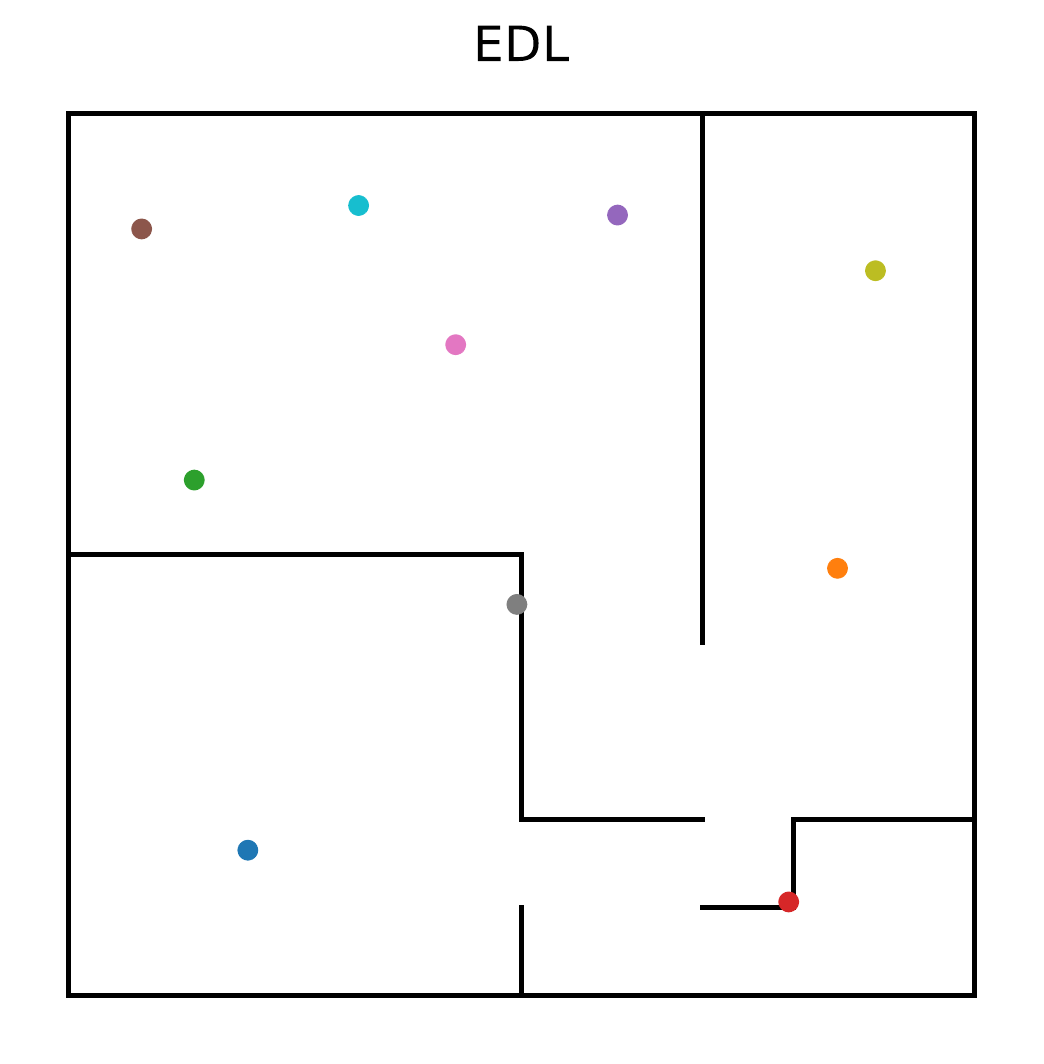}}
    \caption{Goal states discovered by methods using the forward form of the mutual information in Figure~\ref{fig:bottleneck_maze}. We define a goal state as the most likely state under $q_{\phi}(s|z)$ for each skill, i.e.~$g_i = \text{argmax}_s q_{\phi}(s|z_i)$. The baseline method relies on the stationary state-distribution induced by the policy to discover goals. This policy seldom leaves the initial room, limiting the goals that can be discovered. In contrast, the uniform distribution over states in EDL enables the discovery of goals across the whole maze.}
    \label{fig:bottleneck_maze-centroids}
\end{figure}

\begin{figure*}[t]
    \centering
    \resizebox{.8\textwidth}{!}{\includegraphics{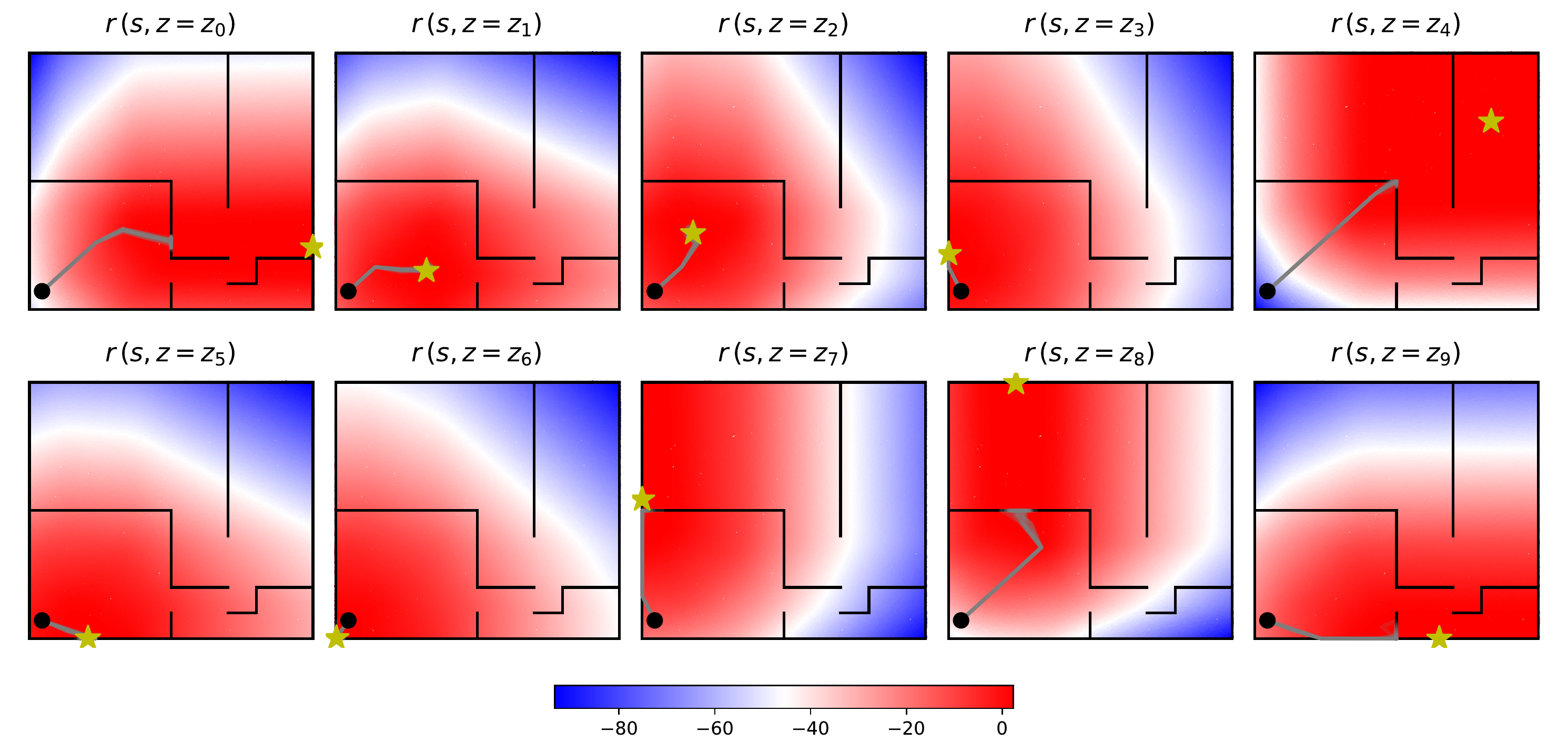}}
    % \resizebox{.8\textwidth}{!}{\includegraphics{figs/forward_mi-reward_landscape_cond-bottleneck_maze.pdf}}
    \caption{Reward landscape per skill at convergence for the agent in Figure~\ref{fig:bottleneck_maze} (left). 
    Trajectories from each skill starting from the black dot are plotted in gray.
    The yellow star indicates the point of maximum reward for each skill. 
    For some skills, this point belongs to an unexplored region of the state space, contrary to the intuition in Section~\ref{sec:theoretical_rew_analysis}. Note that this is due to the Gaussian assumption over $p(s|z)$ in the density model.
    }
    \label{fig:bottleneck_maze-forward_mi_reward_landscape}
\end{figure*}

\begin{figure*}[t]
    \centering
    \resizebox{.8\textwidth}{!}{\includegraphics{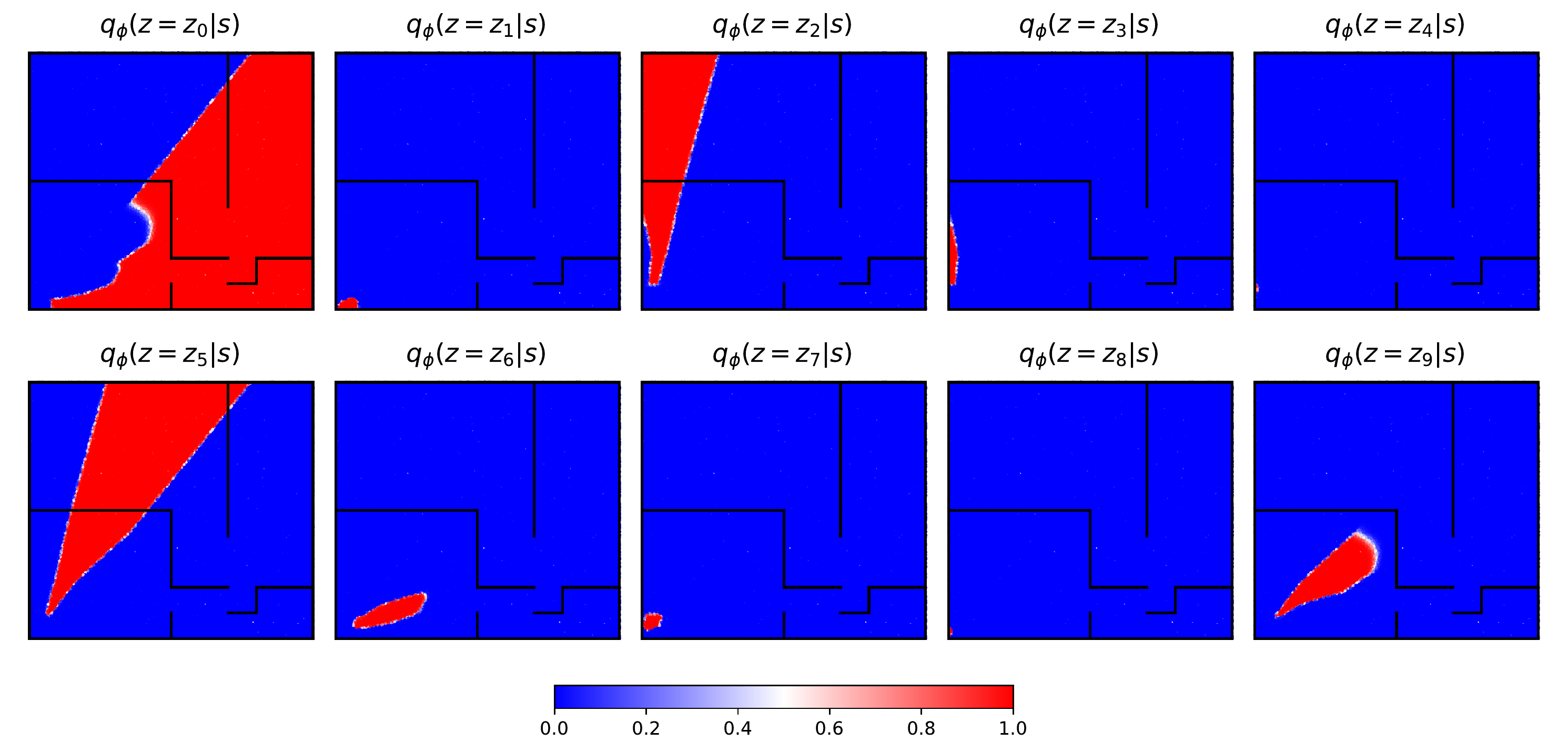}}
    \caption{Approximate posterior $q_{\phi}(z|s)$ at convergence for the agent in Figure~\ref{fig:bottleneck_maze} (middle). Recall that the reward function for this agent is $r(s,z) = \log q_{\phi}(z|s) - \log p(z)$, and $\log p(z)$ is constant in our experiments due to the choice of prior over latent variables. The state space is partitioned in disjoint regions, so that skills only need to enter their corresponding region in order to maximize reward. Note how $q_{\phi}(z|s)$ extrapolates this partition to states that have never been visited by the policy. When combined with an entropy bonus, this reward landscape results in skills that produce highly entropic trajectories within each region.}
    \label{fig:bottleneck_maze-reverse_mi_reward_landscape}
\end{figure*}

\begin{figure*}[t]
    \centering
    \resizebox{.8\textwidth}{!}{\includegraphics{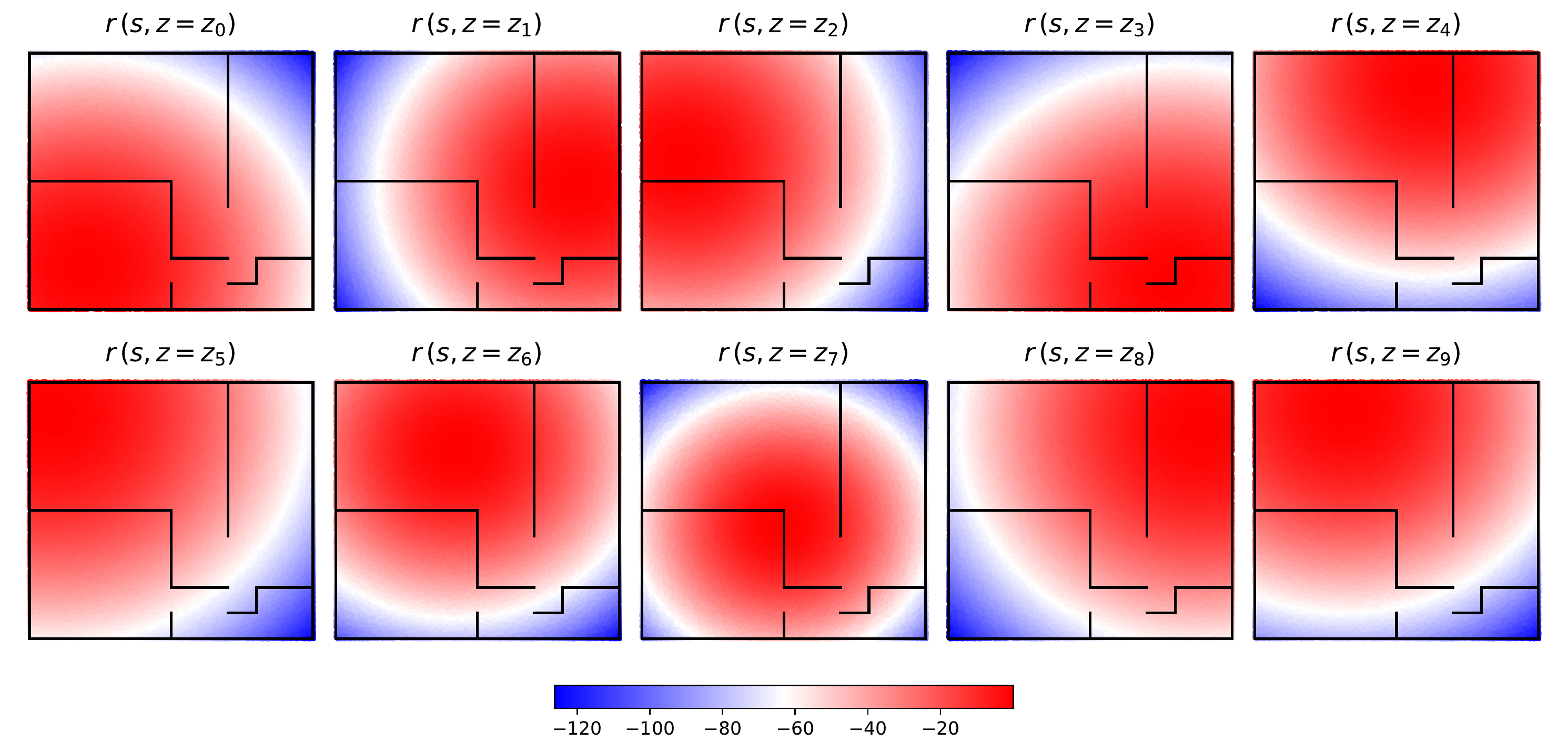}}
    \caption{Reward landscape per skill at convergence for the agent in Figure~\ref{fig:bottleneck_maze} (right). The reward functions follow a bell shape centered at each of the centroids in Figure~\ref{fig:bottleneck_maze-centroids} (right). These are dense signals that ease optimization, but training is prone to falling in local optima due to their deceptive nature.}
    \label{fig:bottleneck_maze-proposed_reward_landscape}
\end{figure*}

\begin{figure*}[t]
    \centering
    \resizebox{1.0\textwidth}{!}{\includegraphics{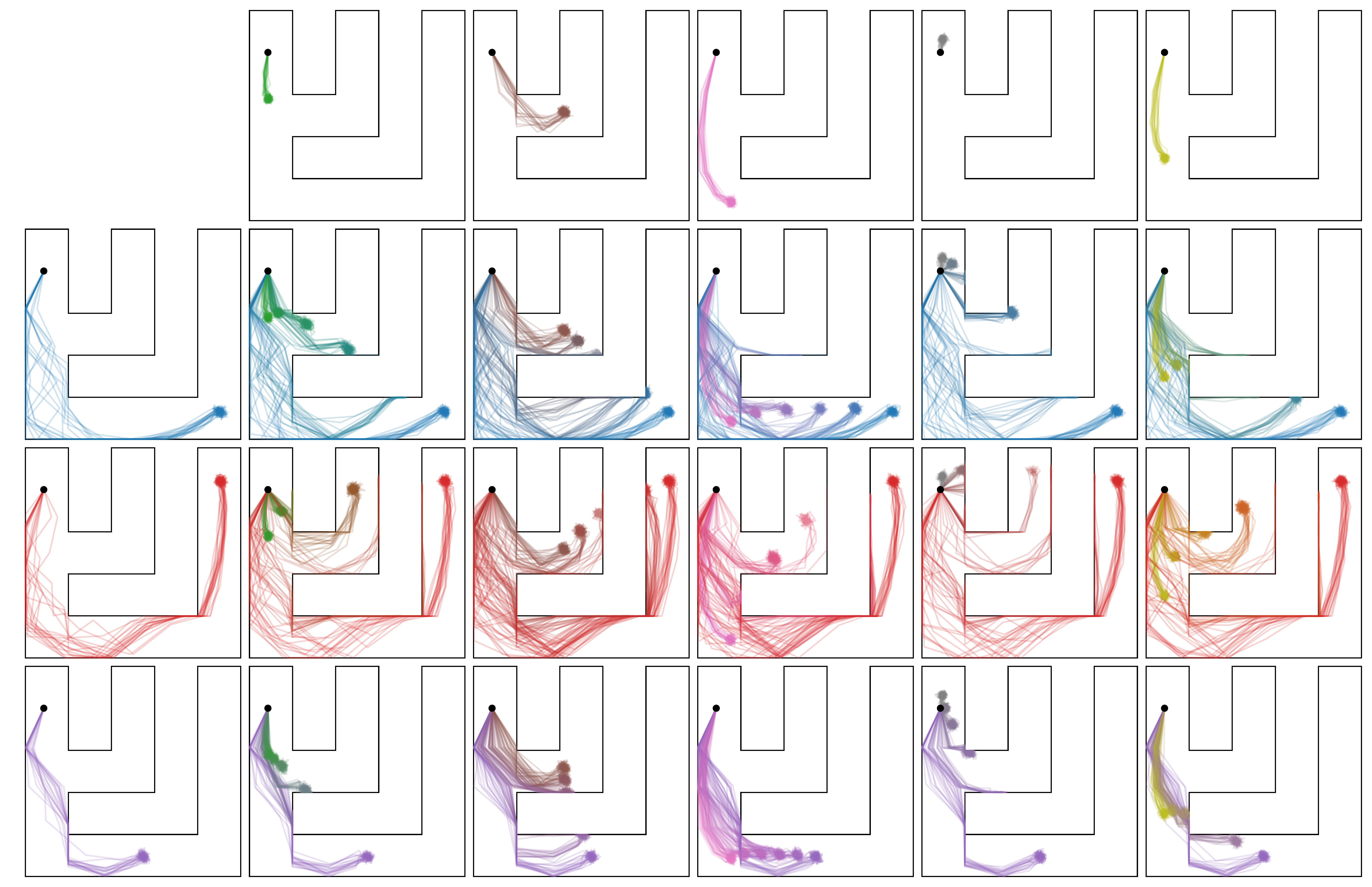}}
    \caption{Interpolating skills learned by EDL. Interpolation is performed at the latent variable level by blending the $z$ vector of two skills. The first row and column show the original skills being interpolated, which were selected randomly from the set of learned options. When plotting interpolated skills, we blend the colors used for the original skills.}
    \label{fig:skill_interpolation_large}
\end{figure*}

We include visualizations that provide further insight about the results presented in the paper, and that could not be included there due to space constraints. These include the goal states discovered by methods using the forward for of the mutual information (Figure~\ref{fig:bottleneck_maze-centroids}), visualization of the reward landscape of each method (Figures~\ref{fig:bottleneck_maze-forward_mi_reward_landscape}, \ref{fig:bottleneck_maze-reverse_mi_reward_landscape} and \ref{fig:bottleneck_maze-proposed_reward_landscape}), and additional skill interpolations (Figure~\ref{fig:skill_interpolation_large}).

% \FloatBarrier
% \bibliography{references}
% \bibliographystyle{icml2020}

%%%%%%%%%%%% END OF APPENDIX %%%%%%%%%%%%%

\end{document}